\documentclass[fleqn,10pt]{wlscirep}
\usepackage[utf8]{inputenc}
\usepackage[T1]{fontenc}
\DeclareUnicodeCharacter{FF0C}{,}
\usepackage{bm}
\usepackage{algorithm}
\usepackage{algorithmic}
\usepackage{lineno} 

\usepackage{amsmath}
\usepackage{amssymb}
\usepackage{mathtools}
\usepackage{amsthm}
\usepackage{bm}
\usepackage{xurl}
\usepackage{amsthm}

\usepackage{ulem}

\title{Learning Design-Score Manifold to Guide Diffusion Models for Offline Optimization}

\author[1,2,3]{Tailin Zhou}
\author[3]{Zhilin Chen}
\author[3]{Wenlong Lyu}
\author[3]{Zhitang Chen}
\author[2,1]{Danny H.K. Tsang}
\author[1,*]{Jun Zhang}
\affil[1]{The Hong Kong University of Science and Technology, Hong Kong, China}
\affil[2]{The Hong Kong University of Science and Technology (Guangzhou), Guangzhou, China}
\affil[3]{Huawei Technologies Co., Ltd.}

\affil[*]{correspondence: Jun Zhang (eejzhang@ust.hk)}

\begin{abstract}
Optimizing complex systems—from discovering therapeutic drugs to designing high-performance materials—remains a fundamental challenge across science and engineering, as the underlying rules are often unknown and costly to evaluate. 
Offline optimization aims to optimize designs for target scores using pre-collected datasets without system interaction.
However, conventional approaches may fail beyond training data, predicting inaccurate scores and generating inferior designs. 
This paper introduces ManGO, a diffusion-based framework that learns the design-score manifold, capturing the design-score interdependencies holistically.
Unlike existing methods that treat design and score spaces in isolation, ManGO unifies forward prediction and backward generation, attaining generalization beyond training data. 
Key to this is its derivative-free guidance for conditional generation, coupled with adaptive inference-time scaling that dynamically optimizes denoising paths. 
Extensive evaluations demonstrate that ManGO outperforms 24 single- and 10 multi-objective optimization methods across diverse domains, including synthetic tasks, robot control, material design, DNA sequence, and real-world engineering optimization.
\end{abstract}
\begin{document}

\flushbottom
\maketitle
\thispagestyle{empty}

 % \linenumbers 

Across scientific and industrial domains, from drug discovery~\cite{jumper2021highly} to engineering superconducting materials~\cite{pogue2023closed}, researchers face a common bottleneck:  creating new designs to optimize specific property scores in complex systems where the rules governing performance are unknown or costly to evaluate. 
Numerous methods require either expensive trial-and-error experimentation or building system models with long-term accumulation of domain knowledge.
Consider the decades-long quest for fusion reactor materials—each physical test costs millions and risks equipment damage~\cite{jha2019enhancing} or the ethical constraints in developing neuroactive drugs where failed designs could harm patients~\cite{drug,stahl2024rethinking}.
These limitations call for a general optimization framework that learns directly from historical data while eliminating the need for iterative evaluation.

Offline optimization~\cite{design-bench}, also named offline model-based optimization, has emerged as a promising solution, enabling design improvement using pre-collected datasets.
 This approach has proven valuable in molecule generation~\cite{wang2021multi}, protein properties~\cite{StantonMGMDGW22}, and hardware accelerators~\cite{KumarYHSL22}. 
 Current methods adopt a unidirectional strategy: 
(i) Training surrogate models to predict scores from designs. 
These predicted scores are then utilized by various optimizers to identify the optimal designs (forward modeling)~\cite{cma-es, bdi};
(ii) Generating designs that are conditioned on desired scores through the use of generative models (backward generation). 
An example of this approach is the employment of diffusion models~\cite{ddom}, wherein small amounts of noise are incrementally added to a design sample, and a neural network is trained to reverse this noise-adding procedure.
However, forward methods mislead optimizers with overconfident out-of-distribution predictions~\cite{design-bench}, while backward methods struggle with out-of-distribution generation (OOG) on unseen conditions \cite{YangGXZWCW24,lei2026boosting}.
These struggles stem from a deeper oversight: existing works operate in isolated design or score spaces, missing the underlying design-score manifold where optimal designs reside.

\begin{figure}[t!]
    \centering
        \includegraphics[width=\linewidth]{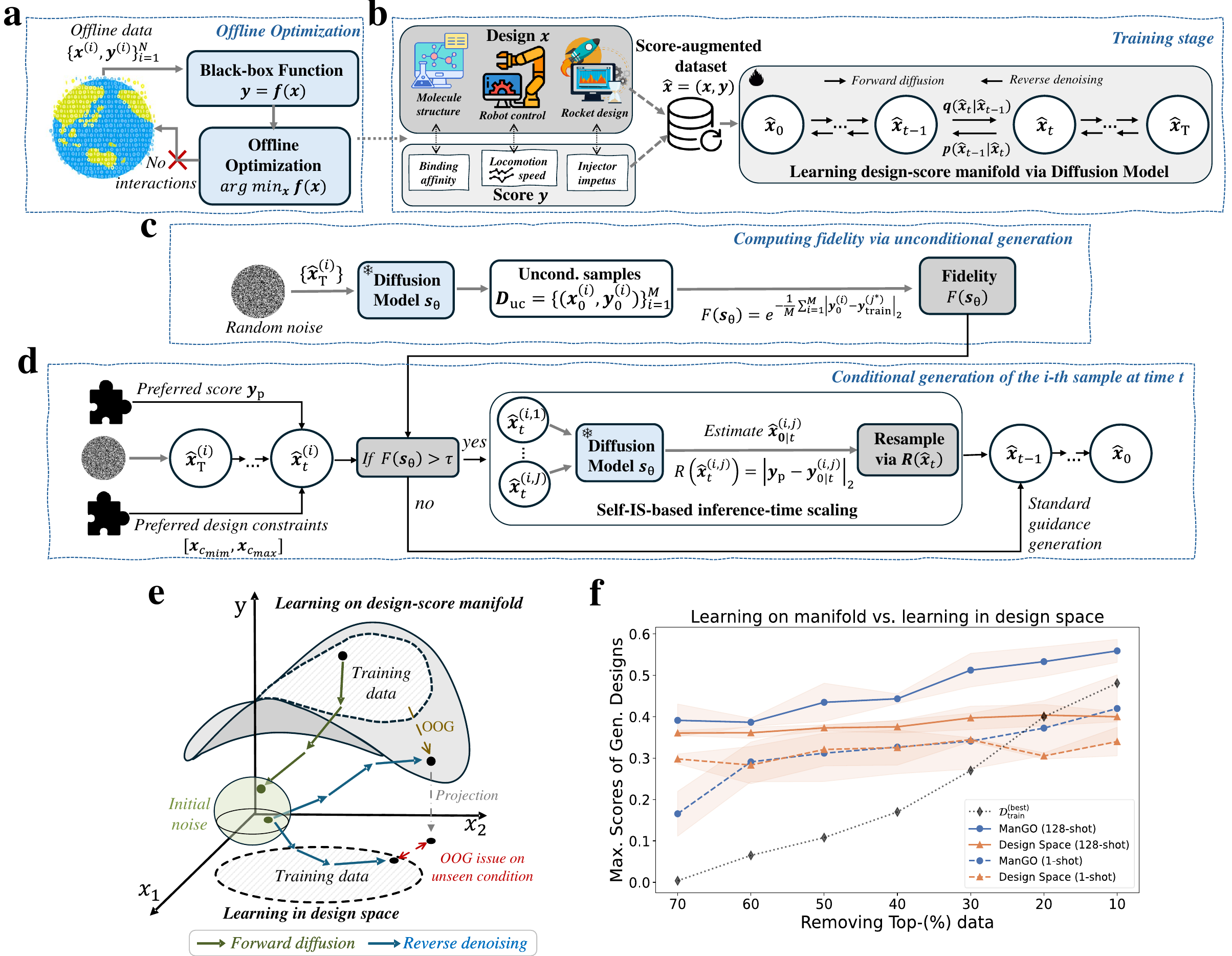}
        % \vspace{-20pt}
    \caption{
    \textbf{Overview of the ManGO framework for offline optimization.}
    (\textbf{a}) Illustration of offline optimization: it identifies optimal designs for an unknown black-box function using an offline dataset (no environment interaction), where designs represent function inputs and scores correspond to outputs.
    (\textbf{b}) Training a diffusion model on score-augmented data to learn the joint design-score manifold.
    (\textbf{c}) Fidelity estimation via unconditional samples generated by the trained ManGO model: the fidelity metric determines whether to activate inference-time scaling during conditional generation.
    (\textbf{d})  Bidirectional conditional generation: it leverages preferred-score or preferred-design conditions to generate corresponding designs or scores, illustrated via the self-supervised importance sampling (self-IS) method at denoising timestep  $t$ for sample $i$.
    (\textbf{e}) Conceptual illustration of ManGO: it learns on the design-score manifold to enhance out-of-distribution generation (OOG) capability, contrasted with design-space learning  that struggles with OOG issues under unseen conditions \cite{YangGXZWCW24}.
    (\textbf{f}) Case study on \textit{superconductor's temperature optimization}~\cite{superconductor}: it demonstrates the superior OOG performance via ManGO versus the design-space approach (i.e., DDOM) across varying ratios of top data removal.
    }
    \label{fig:illustration_figure}
\end{figure}

We propose to learn the design-score \underline{man}ifold to \underline{g}uide diffusion models for offline \underline{o}ptimization (ManGO). 
As illustrated in Figure \ref{fig:illustration_figure}, we leverage score-augmented datasets to train an unconditional diffusion model.  
By learning on the manifold, the model effectively captures the bidirectional relationships between designs and their corresponding scores.
We then introduce a derivative-free guidance mechanism for conditional generation, which enables bidirectional guidance—generating designs based on target scores and predicting scores for given designs, thus eliminating the reliance on error-prone forward models. 
To further enhance generation quality, we implement adaptive inference-time scaling for ManGO, which computes model fidelity on unconditional samples. This scaling approach dynamically optimizes denoising paths through ManGO's self-supervised rewards.
Moreover, ManGO is adaptable to both single-objective optimization (SOO) and multi-objective optimization (MOO) tasks, making it a comprehensive solution for various optimization scenarios.
 
The comprehensive evaluation showcases ManGO's state-of-the-art versatility in offline optimization, consistently surpassing existing approaches across both SOO and MOO tasks.
In comparison with the baselines, ManGO significantly enhances performance, attaining the top position among 24 SOO methods and 10 MOO methods.
We anticipate that ManGO will serve as a valuable tool for data-driven design. By unifying optimization through learning on the design-score manifold, it offers a scalable and accessible solution for future scientific and industrial challenges.

\section*{Results}
In this section, we first introduce preliminaries on offline optimization, the basics of diffusion models, baseline methods and performance metrics.
 Subsequently,  to show the motivation and advantages of ManGO, we compare the learned versus the original design-score manifold and visualize the trajectory generation.  
We then conduct extensive experimental validation on offline SOO and MOO using Design-Bench~\cite{design-bench} and Off-MOO-Bench~\cite{offline-moo}.
Finally, we perform systematic ablation studies to analyze the contributions of  ManGO's core components.

\subsection*{Preliminaries} %Notations and 
% introduce offline optimization
% \subsubsection*{Offline Optimization}
Offline optimization~\cite{design-bench}, also referred to as offline model-based optimization, seeks to identify an optimal design $\bm{x}^*$ within a design space $\mathcal{X} \subseteq \mathbb{R}^d$ without requiring online evaluations, where $d$ denotes the design dimension.  
Based on the number of objective functions $\bm{f}(\cdot) = (f_1(\bm{x}), \dots, f_m(\bm{x})): \mathcal{X} \rightarrow \mathbb{R}^m$, offline optimization can be classified into two types~\cite{offline-moo}: (i) offline SOO when $m = 1$, and (ii) offline MOO when $m > 1$.

%introduce offlineSOO
Offline SOO aims to identify the optimal design $\bm{x}^* = \arg \min_{\bm{x} \in \mathcal{X}} f(\bm{x})$ using only a pre-collected offline dataset $\mathcal{D} = \{(\bm{x}_i, y_i)\}_{i=1}^N$, where $\bm{x}_i$ denotes a specific design (also referred to as a solution) and $y_i = f(\bm{x}_i)$ represents its corresponding score (or objective value).
%introduce offlineMOO
Offline MOO aims to identify a set of designs that achieve optimal trade-offs among conflicting objectives using a pre-collected dataset $\mathcal{D} = \left\{(\bm{x}_i, \bm{y}_i)\right\}_{i=1}^N$, where $\bm{y}_i$ denotes the vector of scores corresponding to design $\bm{x}_i$.  
The problem is defined as~\cite{mo-book}:  
$
\text{Find } \bm{x}^* \in \mathcal{X} \text{ such that } \nexists \bm{x} \in \mathcal{X} \text{ with } \bm{f}(\bm{x}) \prec \bm{f}(\bm{x}^*),  
$
where $\prec$ denotes Pareto dominance.  
A design $\bm{x}^\prime$ is said to \textit{Pareto dominate} another design $\bm{x}$, denoted as $\bm{f}(\bm{x}^\prime) \prec \bm{f}(\bm{x})$, if 
$
\exists i \in \{1, \ldots, m\}, f_i(\bm{x}^\prime) < f_i(\bm{x})$
and 
$
\forall j \in \{1, \ldots, m\}, f_j(\bm{x}^\prime) \leq f_j(\bm{x}).  
$  
Namely, $\bm{x}^\prime$ is superior to $\bm{x}$ in at least one objective while being at least as good in all others.
A design $\bm{x}^*$ is \textit{Pareto optimal} if no other design $\bm{x} \in \mathcal{X}$ Pareto dominates $\bm{x}^*$.  
The set of all Pareto optimal designs is referred to as the \textit{Pareto set (PS)}, and the set of their scores $\{\boldsymbol{f}(\bm{x}^*) \mid \bm{x}^* \in \text{PS}\}$ constitutes the \textit{Pareto front}.
The goal of offline MOO is to identify the PS using a pre-collected dataset, thereby achieving optimal trade-offs among conflicting objectives.

Diffusion models are a type of deep generative models that learn to reverse a gradual noising process, transforming random noise into realistic data through iterative denoising.
Let $\bm{x}_t$ denote the state of a data sample $\bm{x}_0$ at time $t \in [0, T]$, where $\bm{x}_0$ is drawn from an unknown data distribution $p_0(\bm{x})$.
Here, $\bm{x}_t$ represents a noisy version of $\bm{x}_0$ at time $t$, and $\bm{x}_T$ corresponds to a point sampled from a prior noise distribution $p_T(\bm{x})$, typically chosen as the standard normal distribution $p_T(\bm{x}) = \mathcal{N}(\bm{0}, \bm{I})$.

The forward diffusion process, also known as the noise-adding process, can be modeled as a stochastic differential equation (SDE)~\cite{sde_diffusion}:
$
% \begin{equation}
% \label{eq:diffusion_forward}
    d\bm{x} = \mathbf{f}(\bm{x}, t) dt + g(t) d\bm{w},
% \end{equation}
$
where $\bm{w}$ denotes the standard Wiener process, $\mathbf{f}: \mathbb{R}^d \rightarrow \mathbb{R}^d$ is the drift coefficient, and $g(t): \mathbb{R} \rightarrow \mathbb{R}$ is the diffusion coefficient of $\bm{x}_t$.
The denoising process is defined by the reverse-time SDE:
$
% \begin{equation}\label{eq:diffusion_reverse}
    d\bm{x} = \left[\mathbf{f}(\bm{x}, t) - g(t)^2 \nabla_{\bm{x}} \log p_t(\bm{x})\right] dt + g(t) d\tilde{\bm{w}},
% \end{equation}
$
where $dt$ represents an infinitesimal step backward in time, and $d\tilde{\bm{w}}$ is the reverse-time Wiener process.
% This work follows continuous-time diffusion models ~\cite{ddpm} for offline optimization tasks.

\subsection*{Baseline Methods and Performance Metrics} 
We consider existing baseline methods for offline SOO based on three methodological paradigms:
(i) \textit{Surrogate-based methods}: optimizing with surrogate models, including BO-$q$EI~\cite{bo-book, bo-tutorial}, CMA-ES~\cite{cma-es}, REINFORCE~\cite{reinforce}, Gradient Ascent and its variants of mean ensemble and min ensemble. 
(ii) \textit{Forward-modeling methods}: employing advanced neural networks like generative models as surrogate models and integrating with surrogate-based methods, including COMs~\cite{coms}, RoMA~\cite{roma}, IOM~\cite{iom}, BDI~\cite{bdi}, ICT~\cite{ict}, Tri-Mentoring~\cite{tri-mentoring}, PGS~\cite{pgs}, FGM~\cite{fgm}, Match-OPT~\cite{match-opt}, and RaM~\cite{tan2024offline}.
(iii)  \textit{Inverse-modeling methods}: applying score as a condition to reverse design with generative models, including CbAS~\cite{cbas}, MINs~\cite{mins}, DDOM~\cite{ddom}, BONET~\cite{bonet}, and GTG~\cite{gtg}.

For offline MOO, existing approaches remain relatively under-explored compared to offline SOO. 
Our evaluation focuses on three representative approaches:
(i) \textit{Multiple Models (MMs)-based} NSGA-II: 
We implement NSGA-II with independent objective predictors and perform predictors' ensemble as the surrogate model for evolutionary optimization, which outperforms end-to-end and multi-head variants \cite{offline-moo}.
(ii) \textit{Multi-objective Bayesian Optimization} (MOBO): 
We adapt the canonical MOBO by substituting Gaussian Processes with the MM ensemble and employ an HV-based acquisition function, qNEHVI~\cite{daulton2021parallel}, which outperforms scalarization and information-theoretic alternatives \cite{offline-moo}.
(iii)  \textit{Generative methods}: 
ParetoFlow~\cite{yuan2024paretoflow},  a flow-model-based method utilizing adaptive weights for multiple predictors to guide flow sampling toward PF; 
MO-DDOM, a diffusion-model-based method where we extend DDOM through multi-score conditioning and adding MM-based design evaluation.

In offline optimization, where environment interaction is prohibited, it is essential to evaluate multiple candidate solutions; thus, standard benchmarks adopt $k$-shot evaluation with the 100th percentile (best candidate) as the performance metric \cite{design-bench,offline-moo}.
All results are normalized using task-specific references for comparison:
For Design-Bench with maximization tasks (Table \ref{tab:results-designben}), we use standardization normalization for the Superconduct task and min-max normalization for other tasks based on the unobserved dataset's highest score.
For the Off-MOO-Bench with minimization tasks (Tables \ref{tab:results_moo_synthetic_avg}), we use min-max normalization with the best HV and IGD values of training datasets.

  \begin{figure}[t!]
    \centering
       \includegraphics[width=0.6\linewidth]{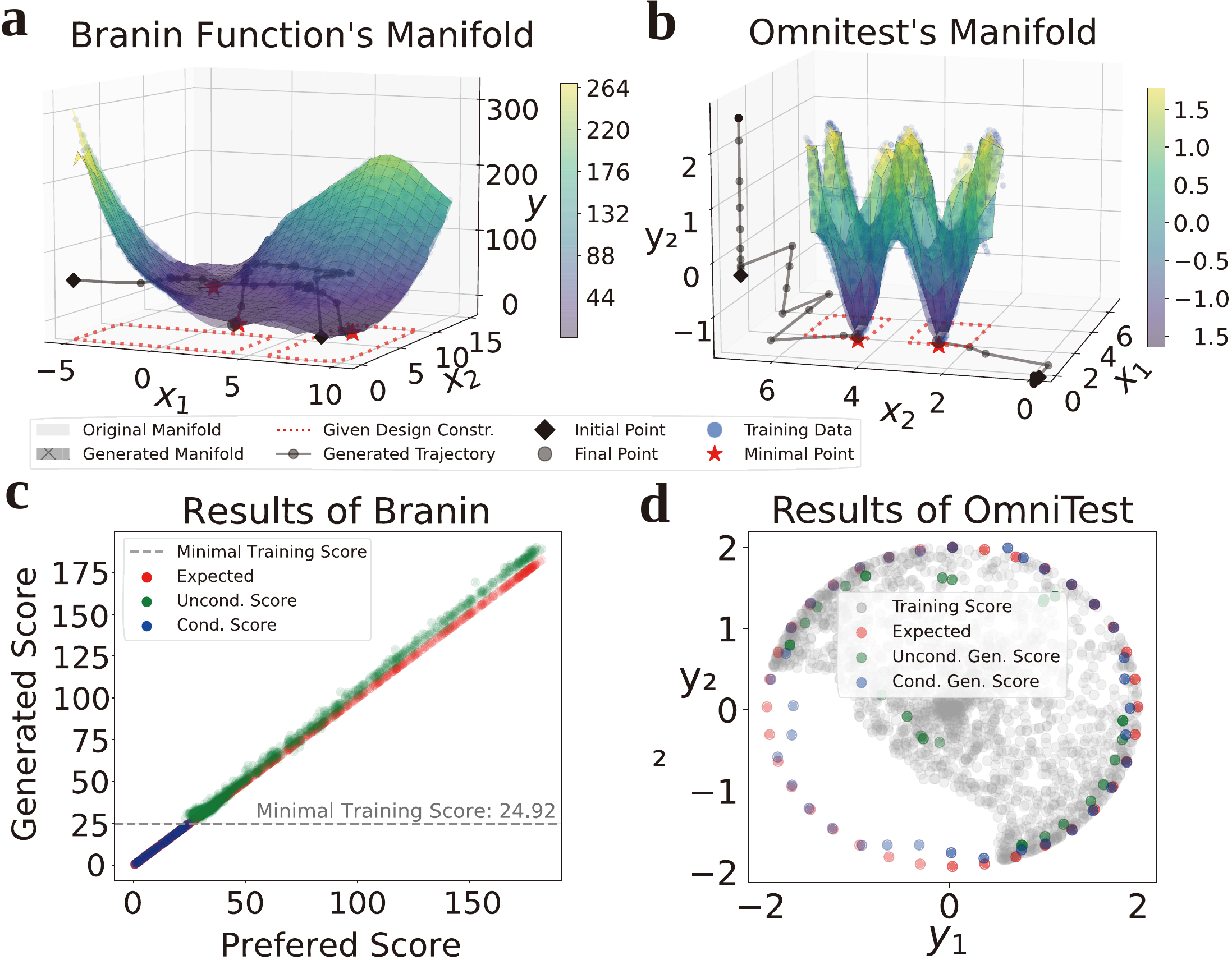}
    \caption{
\textbf{Visualization of manifold learning, trajectory generation, and generation capabilities of ManGO.}
Note that unconditional and conditional samples are generated via ManGO without guidance and with preferred-score guidance, respectively.
(\textbf{a-b}) Manifold and trajectory comparisons for the Branin (SOO) and OmniTest (MOO) tasks.
 The generated manifold is constructed via ManGO’s design-to-score prediction within the feasible region of designs.
Close alignment between the ManGO-generated and original manifold, confirming the model’s proficiency in learning complex design-score relationships. 
Generated trajectories visualize ManGO’s score-to-design mapping under minimal score and design constraints, highlighting its capacity to perform targeted denoising toward desired regions.
(\textbf{c}) Branin task: Unconditional samples (green) match preferred scores from the training dataset, while conditional samples (blue) extrapolate beyond the training minimum (grey dashed line).
(\textbf{d}) OmniTest task: Conditional samples better approximate preferred scores and Pareto-dominate the training data (grey) compared to unconditional samples.
These results indicate that ManGO effectively reconstructs in-distribution samples during unconditional generation—reflecting well-learned manifold structure—while enabling OOG of superior samples through conditional guidance, demonstrating robust extrapolation based on the learned manifold.
}
  \label{fig:manifold_soo_moo_row}
\end{figure}

\begin{figure}[t!]
    \centering
    \includegraphics[width=0.9\linewidth]{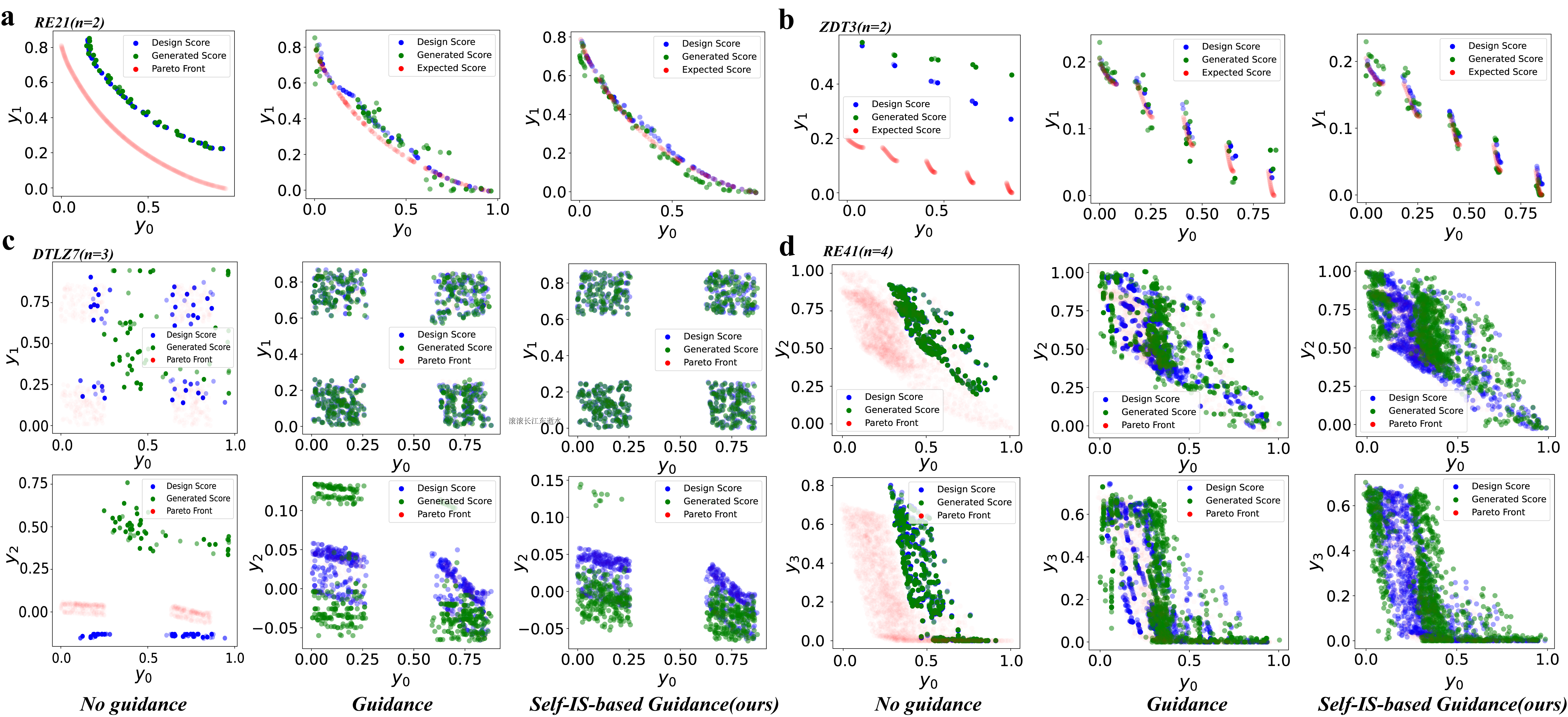}
    \caption{
\textbf{Pareto front generation under different guidance conditions.}
Across all subfigures of (\textbf{a}) RE21, (\textbf{b})  ZDT3,  (\textbf{c}) DTLZ7, and (\textbf{d})  RE41, columns from left to right respectively show the results of no guidance,   standard guidance, and the proposed self-IS-based guidance. 
The progressive improvement in generation quality highlights ManGO’s capability in OOG under conditional guidance. 
Furthermore, the enhanced performance with self-IS-based guidance illustrates the ManGO’s feasibility for more delicate guidance mechanisms.
}
    \label{fig:combined_vis}
\end{figure}

 \subsection*{Motivation and Advantages of ManGO}

Compared to conventional manifold learning, our diffusion-based approach provides unique advantages for offline optimization.
Unlike conventional methods like kernel-based methods that struggle with complex nonlinear geometries \cite{meilua2024manifold}, diffusion models excel at capturing intricate manifold structures through stochastic denoising.
Crucially, diffusion models inherently support conditional generation, enabling direct generation of high-performing designs conditioned on target scores–a capability absent in standard manifold learning.
Furthermore, compared to GANs or VAEs, diffusion models offer superior training stability and generation fidelity \cite{ddpm}, which are critical for reliable optimization from offline data. 
This combination of strong representational capacity and built-in conditional generation makes diffusion models suited for learning the design-score manifold.

Compared to existing methods for offline optimization, our proposed ManGO framework explicitly learns the design-score manifold and leverages the underlying manifold geometry to co-generate designs and scores.
Specifically,  ManGO learns a bidirectional generation between designs and scores:
(i) \textit{Design-to-Score Prediction}: Given any design configuration, ManGO predicts its corresponding score;
(ii)  \textit{Score-to-Design Generation}: For any preferred score, ManGO generates its corresponding design.
As illustrated in Figure \ref{fig:illustration_figure}.e, this bidirectional mapping provides ManGO with a robust OOG capability to extrapolate beyond training distributions.
Let $\hat{\bm{x}}_t=({\bm{x}}_t,{\bm{y}}_t)$ denotes a score-augmented design vector at timestep $t$, where $\bm{x}$  and $\bm{y}$  represent the design and its score, respectively.
The denoising update can be represented as:

$$
\hat{\bm{x}}_{t - 1} = ({\bm{x}}_{t}  ,{\bm{y}}_{t} )+  \underbrace{(\Delta {\bm{x}}_{t}, \Delta {\bm{y}}_{t})}_{\text{Co-update}}=\frac{1}{2}\underbrace{({\bm{x}}_{t}+2\Delta {\bm{x}}_{t}, {\bm{y}}_{t})}_{\text{Design update}} + \frac{1}{2}\underbrace{({\bm{x}}_{t}, {\bm{y}}_{t}+2\Delta {\bm{y}}_{t})}_{\text{Score update}},
$$ 
where the co-update term indicates the denoising update based on the learned manifold geometry, the design-update term indicates the denoising update of ${\bm{x}}_{t}$  conditioned on ${\bm{y}}_{t}$, and the score-update term indicates the denoising update of ${\bm{y}}_{t}$  conditioned on ${\bm{x}}_{t}$.
Unlike the design-space approaches that treat $\bm{y}_{t}$ as a fixed condition and only update $\bm{x}_{t}$  unidirectionally, ManGO jointly updates both  $\bm{x}_{t}$ and $\bm{y}_{t}$ based on the bidirectional mapping.
The design-space methods struggle with extrapolation under unseen conditions.
 In contrast, ManGO leverages co-updates to dynamically capture the geometric relationship between design and score:
each denoising step not only pushes $\bm{x}_{t}$ toward the local manifold conditioned on $\bm{y}_{t}$ (design-update) but also refines $\bm{y}_{t}$ to align with the evolving $\bm{x}_{t}$  (score-update).
 This bidirectional feedback enables progressive extrapolation and converges to the conditioned points on the manifold,  achieving a robust OOG capability.

We conduct a controlled experiment to elucidate the advantage of the bidirectional mapping of ManGO on a \textit{superconductor}~\cite{superconductor} task, an 86-D materials design optimization task to maximize a critical temperature, using DDOM~\cite{ddom} as the design-space baseline.
In terms of the 128-shot evaluation, Figure \ref{fig:illustration_figure}.f shows that ManGO achieves a consistent score gain of greater than 0.1 over the design-space approach across varying levels of top-data removal (from 70\% to 10\%).  
The gain grows to nearly 0.2 when only 10\% of the top data are removed.
Regarding the 1-shot evaluation, both methods exhibit comparable performance under severe data removal (from 70\% to 30\%).
However,  ManGO demonstrates progressively better results as data availability increases.
At 10\% data removal,  ManGO achieves higher scores than the design-space method with 128 shots.
These results confirm that: 
(i) ManGO captures the bidirectional design-score relationships, enabling robustness to OOG challenges;
(ii) ManGO exhibits nonlinear scaling of sample efficiency with data quality, achieving superior few-shot performance.

 \subsection*{Manifold and Trajectory Generation Visualization}

To demonstrate the bidirectional mapping capability, we visualize the performance of ManGO on two canonical minimization tasks.
(i) \textit{Branin function} (for SOO): A well-studied 2D function containing three global minima within  $x_1 \in[-5,10]$, $x_2 \in[0,15]$ and $y_{\rm min}=0.398$,  serving as an ideal testbed to capture multimodal landscapes.
Specifically,
$f_{\rm br}\left(x_1, x_2\right)= a\left(x_2-b x_1^2+c x_1-r\right)^2-s(1-t) \cos x_1-s,$
where $a=1, b=\frac{5.1}{4 \pi^2}, c=\frac{5}{\pi}, r=6, s=10$, and $t=\frac{1}{8 \pi}$. 
% In this square region, $f_{b r}$ has three global maximas, $(-\pi, 12.275),(\pi, 2.275)$, and ( $9.42478,2.475)$; with the maximum value of -0.397887.
(ii) \textit{OmniTest} (for MOO): A synthetic 2D problem generating 9 disconnected Pareto-optimal points with  $\bm{x} \in [0,6]^2$ and $\bm{y} \in [-2,2]^2$, challenging optimization methods in maintaining diverse designs.
Specifically,
$ f_1(\bm{x}) = \sum_{i=1}^2 \sin(\pi x_i), f_2(\bm{x}) = \sum_{i=1}^2 \cos(\pi x_i)$  
with  Pareto designs at all combinations of $(x_1, x_2) \in \{1,3,5\}\times\{1,3,5\}$.
% forming 9 disconnected regions in the objective space $[-2,2]^2$ with ideal point $(-2,-2)$ and nadir point $(2,2)$.

%manifold comparison
Figure \ref{fig:manifold_soo_moo_row} shows that ManGO reconstructs the entire design-score manifold despite removing the top 40\%  of low-scoring data.
Its generated manifold recovers the erased global minima locations and maintains the overall topographic trends in the Branin task in Figure \ref{fig:manifold_soo_moo_row}.a.
ManGO also recovers all 9 disconnected Pareto fronts and preserves the negative correlation between $f_1$ and $f_2$ objectives in the  OmniTest task in Figure \ref{fig:manifold_soo_moo_row}.b  (visualizing two regions for ease of observation).
Across both tasks, the generated manifolds exhibit only minimal deviations from the original manifolds, even in out-of-distribution regions. 
This demonstrates ManGO's robust OOG capability, validating its ability to extrapolate beyond the training distribution accurately.
Furthermore,  both generated manifolds exhibit consistent minor elevation, with a deviation of less than 5\%.
For example, Figure \ref{fig:manifold_soo_moo_row}.c reveals a gradual score elevation between unconditional and expected scores in Branin's training region, where higher expected scores correspond to sparser training samples.   
This conservative estimation under uncertainty serves as an advantage in offline optimization, where reliable performance outweighs aggressive extrapolation \cite{coms}.

% Trajectory demonstration
Unlike design-space approaches limited to score-based guidance, ManGO's manifold learning framework enables additional conditioning on design constraints, providing more flexible control over the generation process.
 Figures \ref{fig:manifold_soo_moo_row}.a-b present  ManGO's generated trajectories with minimal score and varying design constraints as conditional guidance. 
 For instance, in Figure \ref{fig:manifold_soo_moo_row}.a with design constraints $x_1 \in [-5,0], x_2 \in[0,15]$ and minimum score condition $y_{\rm min}=0.398$, ManGO successfully guides a randomly initialized point (violating the constraints) to converge to the constrained minimum.
Notably, ManGO exhibits accelerated convergence as noisy samples approach preferred points. 
This indicates its ability to exploit favorable noise points for enhanced output quality, naturally aligning with our inference-time scaling framework.
On the other hand, ManGO directly transports samples from random initializations to preferred points in the joint design-score space, simultaneously generating both designs and their scores.
This eliminates two key requirements of conventional approaches: (i) iterative score evaluation on noisy designs via external forward models, and (ii) gradient computation along manifold geometry.

%conditional and unconditional generation comparison 
Figures \ref{fig:manifold_soo_moo_row} and \ref{fig:combined_vis} demonstrate the critical role of conditional guidance on ManGO's OOG capability.
Unconditional generation faithfully reproduces in-distribution samples (matching training score ranges in Figures \ref{fig:manifold_soo_moo_row}.c-d), while conditional generation produces designs that extrapolate beyond the training distribution. 
This key distinction reveals that although ManGO learns the complete manifold structure, explicit guidance is essential to unlock its full OOG capabilities. 
The consistent results across RE21, ZDT3, DTLZ7, and RE41 benchmarks (Figure \ref{fig:combined_vis}) robustly confirm this fundamental behavior.
Specifically, ManGO without guidance exhibits conservative behavior, remaining within the training distribution.
Meanwhile, when guided by Pareto front (PF) reference points, ManGO approaches the complete PF, and our self-supervised scaling guidance achieves better precision than standard guidance.

 \subsection*{Evaluation on Single-Objective Optimization}
 
 We employ five representative tasks from Design-Bench~\cite{design-bench} and sample $10{,}000$ offline design samples per task ~\cite{tan2024offline}: 
(i) \textit{Ant Morphology}~\cite{gym} (60-D parameter optimization for quadruped locomotion speed),
(ii) \textit{D'Kitty Morphology}~\cite{robel} (56-D parameter optimization for movement efficiency enhancement of a quadruped robot), 
and (iii) \textit{Superconductor}~\cite{superconductor},
 % (86-D material design optimization for critical temperature maximization)
(iv) \textit{TF-Bind-8}~\cite{tfbind} and  (v) \textit{TF-Bind-10}~\cite{tfbind} (discrete DNA sequence optimization for transcription factor binding affinity with sequence lengths 8 and 10). 
We follow the maximization setting of Design-Bench and normalize scores based on the maximal score in the unobserved dataset \cite{design-bench}, where higher scores indicate better performance.

As shown in Table~\ref{tab:results-designben}, we compare ManGO with 22 baseline methods and report normalized scores of top $k=128$ candidates (100th percentile).
ManGO establishes a state-of-the-art performance across diverse domains (materials, robotics, bioengineering) on five datasets. 
ManGO with standard guidance attains a mean rank of 2.2/24 (securing the second position), while the self-supervised importance sampling (self-IS)-based variant further improves this to 1.4/24 (the first). 
 ManGO ranks first on four tasks, including D’Kitty, Superconductor, TF-Bind-8, and TF-Bind-10, and secures second place on Ant, trailing only CMA-ES.
This cross-domain advantage suggests that the effectiveness of learning the design-score manifold is general and not limited to specific problems.

The 13.6-rank leap over Mins (rank 15.0, the previous best of inverse-modeling baselines) demonstrates the superiority of the manifold-learned generation to design-space-learned methods.
Meanwhile, the 2.8-rank lead over RaM (the best of forward-modeling baselines) suggests that score-conditioned diffusion can better exploit offline data than ranking-based approaches.
It also outperforms the top surrogate-based methods (CMA-ES, rank 13.0) by 11.6 ranks, without the need for designing acquisition functions.
On the other hand, the self-IS variant shows consistent improvements over standard ManGO: score boosts on Superconductor (+3.8\%) and modest gains on Ant (+0.8\%) and TF-Bind-10 (+0.9\%).
A deviation occurs in D’Kitty, where a slight average score reduction (-0.2\%) accompanies improved peak performance (+0.3\%).
The marginal gains reflect that standard ManGO reaches near-optimal performance, leaving limited room for improvement.
% The self-IS variant's primary value lies in reliability enhancement, as evidenced by its rank consistency across all tasks.

 \subsection*{Evaluation on  Multi-Objective Optimization}
We utilize Off-MOO-Bench~\cite{offline-moo} and sample $60{,}000$ samples  per task ~\cite{yuan2024paretoflow}: 
(i) \textit{Synthetic Functions} (an established collection of MOO evaluation tasks with $2-3$ objectives exhibiting diverse PF characteristics,  such as ZDT~\cite{zdt} and DTLZ~\cite{dtlz}),
and (ii) \textit{real-world engineering (RE) applications}~\cite{tanabe2020easy} (a suite of practical design tasks with $2-4$ competing objectives,  such as four-bar truss design and rocket injector design).
We employ two standard evaluation metrics: 
(i) \textit{Hypervolume (HV)}~\cite{hv}, which quantifies the dominated volume between candidate designs and nadir point (each dimension of which corresponds to the worst value of one objective), 
and (ii) \textit{Inverted Generational Distance (IGD)}~\cite{igd}, which measures the average minimum distance between candidate designs and the ground-true PF,   both metrics applied to non-dominated sorting~\cite{deb2002fast} with $k=256$ candidate designs (100th percentile).
While generating high-quality single solutions from purely offline data remains challenging, we also report our method's performance at $k=1$ to demonstrate competitive performance.
 Note that  while we replace the online query in MOBO/NSGA2 with a surrogate forward model for offline adaptation,  performance degradation occurs versus online operation as the surrogate model cannot perfectly emulate environment feedback.
 
 We follow the minimization setting of Off-MOO-Bench and normalize HV (IGD) values based on the best HV (IGD) of the training dataset, where higher HV (lower IGD) indicates better performance.
ManGO outperforms all baseline methods across both synthetic and real-world MOO benchmarks according to Table \ref{tab:results_moo_synthetic_avg}.
Regarding synthetic tasks, the self-IS-based ManGO achieves the best mean rankings of 2.0 (HV) and 1.3 (IGD) out of 10 competing methods, while the standard guidance version follows closely with ranks of 2.7 (HV) and 1.7 (IGD), securing the top two positions.
The superiority extends to RE tasks, where self-IS-based ManGO dominates with average ranks of 1.3 (HV) and 2.0 (IGD), establishing itself as the overall leader.

As presented in the upper part of Table \ref{tab:results_moo_synthetic_avg}, ManGO shows consistent superiority across ZDT, OmniTest, and DTLZ series. 
Under the most challenging OOG scenarios where preferred designs are distant from ZDT's training data, self-IS-based ManGO outperforms the best baselines by 60.1\% in IGD (vs DDOM) and 3.9\% in HV (vs NSGA-2). 
For 1-shot evaluation settings (i.e., $k=1$), ManGO matches the performance of baseline methods requiring 256-shot evaluations.
    This shows that the efficient learning on the design-score manifold enables high-quality guidance generation even with minimal sampling.

As task complexity escalates with increasing objectives, the self-IS-based variant consistently achieves top performance in both HV and IGD metrics in the lower part of Table \ref{tab:results_moo_synthetic_avg}.
This confirms its OOG capability in high-dimensional objective spaces.
Compared to synthetic tasks, learning on manifolds poses greater challenges for RE tasks.
Consequently, this diminishes ManGO's performance advantage in 1-shot and standard guidance modes. 
However, self-IS guidance effectively offsets this by exploring more noise points, and its performance gains become more pronounced as task complexity increases.

\begin{table*}[t!]
\caption{The 100th percentile normalized score ($k=128$) in the Design-Bench benchmark, where the best and runner-up results on each task are \textbf{bold} and \underline{underlined}  numbers. $\mathcal{D}^{\text{(best)}}_{\text{train}}$  denotes the best score in the offline training dataset. Each task's results are normalized by the best scores in the unobserved dataset.
DDOM represents a conditional diffusion-based method. }
\centering
\resizebox{0.8\linewidth}{!}{\begin{tabular}{ccccccc}
\toprule
Method        & Ant                                        & D'Kitty                                    & Superconductor                             & TF-Bind-8                                 & TF-Bind-10                                  & Mean Rank                                         \\  \midrule 
$\mathcal{D}^{\text{(best)}}_{\text{train}}$ (Preferred)  & 0.565 (1.0) & 0.884 (1.0) & 0.400 (1.0)& 0.439 (1.0)& 0.467 (1.0)& /   \\ \midrule
BO-$q$EI            & 0.812 ± 0.000                              & 0.896 ± 0.000                              & 0.382 ± 0.013                              & 0.802 ± 0.081                              & 0.628 ± 0.036                            & 20.0 / 24                             \\
CMA-ES            & \textbf{1.214 ± 0.732}     & 0.725 ± 0.002                              & 0.463 ± 0.042                              & 0.944 ± 0.017                              & 0.641 ± 0.036                              & 13.0 / 24                             \\
REINFORCE         & 0.248 ± 0.039                              & 0.541 ± 0.196                              & 0.478 ± 0.017                              & 0.935 ± 0.049                              & 0.673 ± 0.074                              & 16.0 / 24                             \\
Grad. Ascent      & 0.273 ± 0.023                              & 0.853 ± 0.018                              & 0.510 ± 0.028                              & 0.969 ± 0.021                              & 0.646 ± 0.037                              & 13.6 / 24                             \\
Grad. Ascent Mean & 0.306 ± 0.053                              & 0.875 ± 0.024                              & 0.508 ± 0.019                              & \underline{{0.985 ± 0.008}}   & 0.633 ± 0.030                            & 13.2 / 24                             \\
Grad. Ascent Min  & 0.282 ± 0.033                              & 0.884 ± 0.018                              &  0.514 ± 0.020 & 0.979 ± 0.014                              & 0.632 ± 0.027                                                         & 13.5 / 24                             \\ \midrule
COMs              & 0.916 ± 0.026                              & 0.949 ± 0.016                              & 0.460 ± 0.040                              & 0.953 ± 0.038                              & 0.644 ± 0.052                              & 11.5 / 24                              \\
RoMA              & 0.430 ± 0.048                              & 0.767 ± 0.031                              & 0.494 ± 0.025                              & 0.665 ± 0.000                              & 0.553 ± 0.000                              & 20.3 / 24                            \\
IOM               & 0.889 ± 0.034                              & 0.928 ± 0.008                              & 0.491 ± 0.034                              & 0.925 ± 0.054                              & 0.628 ± 0.036                              & 15.1 / 24                             \\
BDI               &  0.963 ± 0.000  & 0.941 ± 0.000                              & 0.508 ± 0.013                              & 0.973 ± 0.000                              & 0.658 ± 0.000                                                        &  7.7 / 24                              \\
ICT               & 0.915 ± 0.024                              & 0.947 ± 0.009                              & 0.494 ± 0.026                              & 0.897 ± 0.050                              & 0.659 ± 0.024                              & 11.4 / 24                              \\
Tri-Mentoring     & 0.891 ± 0.011                              & 0.947 ± 0.005                              & 0.503 ± 0.013                              & 0.956 ± 0.000                              & 0.662 ± 0.012                              & 9.7 / 24                              \\
PGS               & 0.715 ± 0.046                              & 0.954 ± 0.022 & 0.444 ± 0.020                              & 0.889 ± 0.061                              & 0.634 ± 0.040                                                         &  15.2 / 24                             \\
FGM               & 0.923 ± 0.023                              & 0.944 ± 0.014                              & 0.481 ± 0.024                              & 0.811 ± 0.079                              & 0.611 ± 0.008                              & 15.2 / 24                             \\
Match-OPT         & 0.933 ± 0.016                              & 0.952 ± 0.008                              & 0.504 ± 0.021                              & 0.824 ± 0.067                              & 0.655 ± 0.050                              & 10.0 / 24                              \\  
 RaM-RankCosine    & 0.940 ± 0.028                              & 0.951 ± 0.017                              &  0.514 ± 0.026                               &  0.982 ± 0.012  & \underline{{0.675 ± 0.049}}                       &  4.5 / 24  \\
 RaM-ListNet       & 0.949 ± 0.025                              &  \underline{{0.962 ± 0.015}}    & 0.517 ± 0.029  & 0.981 ± 0.012                              & 0.670 ± 0.035                                                                                &  4.2 / 24    \\ \midrule 
CbAS              & 0.846 ± 0.032                              & 0.896 ± 0.009                              & 0.421 ± 0.049                              & 0.921 ± 0.046                              & 0.630 ± 0.039                              & 17.5 / 24                             \\
MINs              & 0.906 ± 0.024                              & 0.939 ± 0.007                              & 0.464 ± 0.023                              & 0.910 ± 0.051                              & 0.633 ± 0.034                              & 15.0 / 24                            \\
BONET             & 0.921 ± 0.031                              & 0.949 ± 0.016                              & 0.390 ± 0.022                              & 0.798 ± 0.123                              & 0.575 ± 0.039                              & 17.1 / 24                             \\
GTG               & 0.855 ± 0.044                              & 0.942 ± 0.017                              & 0.480 ± 0.055                              & 0.910 ± 0.040                              & 0.619 ± 0.029                              & 15.9 / 24                            \\ 
DDOM              & 0.908 ± 0.024                              & 0.930 ± 0.005                              & 0.452 ± 0.028                              & 0.913 ± 0.047                              & 0.616 ± 0.018                              & 16.6 / 24                             \\ \midrule
\textbf{ManGO}       & 0.960 ± 0.017                          &  \textbf{{0.971 ± 0.004}}    & \underline{{0.523 ± 0.040}}  &  \textbf{{0.985 ± 0.004} }                            & 0.673 ± 0.033                         & \underline{{2.2 / 24}}   \\ 
\textbf{ $\text{ManGO}^\text{+Self-IS}$}       & \underline{{0.968 ± 0.013}}                            &  \underline{{0.969 ± 0.009}}    & \textbf{{0.543 ± 0.037}}   &  \textbf{{0.985 ± 0.004} }                            & \textbf{{0.679 ± 0.023}    }                              & \textbf{{1.4 / 24}}   \\ 
\bottomrule 
\end{tabular}}\label{tab:results-designben}
\end{table*}

\begin{table*}[t]
\centering
\caption{Averaged normalized HV and IGD values of synthetic tasks (upper) and  RE tasks (lower) in the Off-MOO benchmark, where the best and runner-up results on each task are highlighted by \textbf{bold} and \underline{underlined} numbers. 
Higher HV values indicate better performance, while lower IGD values are preferred.
$\mathcal{D}^{\text{(best)}}_{\text{train}}$(Preferred)  denotes the best (preferred) HV/IGD in the offline training dataset.
For compact presentation, reported numbers represent tasks' performance averaged by the objective number $n$.
The sets of ZDT($n$=2),  OmniTest($n$=2), and  DTLZ($n$=3) consist of 5 ZDT tasks \cite{zdt},  1 OmniTest task \cite{omnitest}, and  2 DTLZ tasks \cite{dtlz}, respectively.
The sets of RE($n=2$), RE($n=3$), and RE($n=4$) comprise 5, 7, and 2 real-world application tasks \cite{tanabe2020easy} with $n=2$, $3$, and $4$, respectively.
Note that each task's results are normalized by the best HV and IGD of its training dataset, and RE($n$=2) presents higher averaged IGD values because the RE22 task has ten times more IGD value than other tasks.
Note that  MO-DDOM represents a standard conditional diffusion-based baseline method.
} 
\centering
\resizebox{\linewidth}{!}{
\begin{tabular}{@{}ccccccccc@{}}
\toprule
                                                                    & \multicolumn{2}{c}{ZDT($n$=2)}                  & \multicolumn{2}{c}{OmniTest($n$=2)}             & \multicolumn{2}{c}{DTLZ($n$=3)}                 & \multicolumn{2}{c}{Mean Rank}                \\
                                                                    & Avg. HV($\uparrow$)    & Avg. IGD($\downarrow$) & Avg. HV($\uparrow$)    & Avg. IGD($\downarrow$) & Avg. HV($\uparrow$)    & Avg. IGD($\downarrow$) & HV Rank($\downarrow$) &   IGD Rank($\downarrow$) \\ \midrule
$\mathcal{D}^{\text{(best)}}_{\text{train}}$(Preferred)             & 1.0 (1.118)            & 1.0 (0.0)              & 1.0 (1.056)            & 1.0 (0.0)              & 1.0 (1.098)            & 1.0 (0.0)              & /                   & /                      \\ \midrule
MM-NSGA2($k$=1)                                                     & 0.888 ± 0.013          & 2.555 ± 0.118          & 1.044 ± 0.010          & 0.454 ± 0.129          & 0.987 ± 0.012          & 0.635 ± 0.048          & 8.3 / 10           & 6.7 / 10              \\
MO-DDOM ($k$=1)                                                     & 0.948 ± 0.009          & 1.674 ± 0.150          & 0.983 ± 0.002          & 0.827 ± 0.046          & 0.932 ± 0.014          & 0.795 ± 0.063          & 9.3 / 10           & 8.0 / 10               \\
\textbf{ManGO} ($k$=1)                             & 1.097 ± 0.004          & 0.729 ± 0.064          & 1.050 ± 0.002          & 0.399 ± 0.068          & 0.971 ± 0.016          & 0.739 ± 0.104          & 6.0 / 10            & 6.0 / 10               \\
\textbf{$\text{ManGO}^\text{+Self-IS}$} ($k$=1)    & 1.099 ± 0.005          & 0.702 ± 0.065          & 1.050 ± 0.001          & 0.359 ± 0.017          & 1.053 ± 0.015          & 0.250 ± 0.084          & 4.3 / 10           & 3.7 / 10              \\ \midrule
MM-MOBO                                                             & 0.963 ± 0.007          & 4.723 ± 0.164          & \textbf{1.056 ± 0.000} & 0.206 ± 0.019          & 1.075 ± 0.000          & 0.362 ± 0.016          & 4.0 / 10           & 6.0 / 10               \\
ParetoFlow                                                          & 1.000 ± 0.008          & 2.867 ± 0.405          & 0.953 ± 0.057          & 1.523 ± 0.567          & 0.998 ± 0.009          & 0.672 ± 0.115          & 7.7 / 10           & 8.3 / 10              \\
MM-NSGA2 ($k$=256)                                                  & 1.055 ± 0.003          & 3.592 ± 0.044          & 1.046 ± 0.002          & 1.008 ± 0.019          & \textbf{1.086 ± 0.000} & 0.752 ± 0.016          & 4.0 / 10            & 9.0 / 10               \\
MO-DDOM ($k$=256)                                                   & 0.981 ± 0.006          & 1.052 ± 0.144          & 1.033 ± 0.001          & 0.270 ± 0.010          & 1.054 ± 0.006          & 0.255 ± 0.044          & 6.7 / 10           & 4.3 / 10              \\
\textbf{ManGO} ($k$=256)                           & \textbf{1.107 ± 0.002} & \textbf{0.420 ± 0.030} & 1.051 ± 0.002          & \underline{0.118 ± 0.015}    & 1.066 ± 0.003          & \underline{0.172 ± 0.013}    & \underline{2.7 / 10}     & \underline{1.7 / 10}        \\
\textbf{ $\text{ManGO}^\text{+Self-IS}$} ($k$=256) & \underline{1.106 ± 0.002}    & \underline{0.445 ± 0.043}    & \underline{1.052 ± 0.000}    & \textbf{0.094 ± 0.002} & \underline{1.079 ± 0.003}    & \textbf{0.123 ± 0.033} & \textbf{2.0 / 10}   & \textbf{1.3 / 10}     \\ \midrule
\midrule
                                                                    & \multicolumn{2}{c}{RE($n$=2)}                   & \multicolumn{2}{c}{RE($n$=3)}                   & \multicolumn{2}{c}{RE($n$=4)}                   & \multicolumn{2}{c}{Mean Rank}                \\
                                                                    & Avg. HV($\uparrow$)    & Avg. IGD($\downarrow$) & Avg. HV($\uparrow$)    & Avg. IGD($\downarrow$) & Avg. HV($\uparrow$)    & Avg. IGD($\downarrow$) &  HV Rank($\downarrow$) &   IGD Rank($\downarrow$) \\ \midrule
$\mathcal{D}^{\text{(best)}}_{\text{train}}$(Preferred)             & 1.0 (1.037)            & 1.0 (0.0)              & 1.0 (1.082)            & 1.0 (0.0)              & 1.0 (1.310)            & 1.0 (0.0)              & /                   & /                      \\ \midrule
MM-NSGA2($k$=1)                                                     & 1.016 ± 0.004          & 56.958 ± 12.159        & 0.935 ± 0.007          & 2.979 ± 0.159          & 0.780 ± 0.009          & 1.504 ± 0.060          & 9.3 / 10            & 9.7 / 10               \\
MO-DDOM ($k$=1)                                                     & 1.010 ± 0.012          & 2.261 ± 0.220          & 1.046 ± 0.002          & 0.579 ± 0.020          & 1.058 ± 0.003          & 0.845 ± 0.008          & 8.7 / 10            & 4.7 / 10               \\
\textbf{ManGO} ($k$=1)                             & 1.024 ± 0.004          & 6.809 ± 0.740          & 1.051 ± 0.011          & 0.782 ± 0.083          & 1.234 ± 0.009          & 0.421 ± 0.018          & 5.7 / 10            & 6.0 / 10               \\
\textbf{$\text{ManGO}^\text{+Self-IS}$} ($k$=1)    & 1.022 ± 0.004          & 4.717 ± 2.081          & 1.066 ± 0.005          & 0.588 ± 0.027          & 1.240 ± 0.016          & 0.304 ± 0.020          & 4.7 / 10            & 4.3 / 10               \\ \midrule
MM-MOBO                                                             & 1.027 ± 0.002          & \textbf{1.156 ± 0.133} & \underline{1.071 ± 0.001}    & 0.912 ± 0.093          & 1.123 ± 0.009          & 0.816 ± 0.029          & 3.7 / 10            & 5.0 / 10               \\
ParetoFlow                                                          & 1.017 ± 0.004          & 12.403 ± 0.052         & 1.010 ± 0.004          & 1.806 ± 0.024          & 0.668 ± 0.001          & 1.887 ± 0.006          & 9.0 / 10            & 8.7 / 10               \\
MM-NSGA2 ($k$=256)                                                  & \textbf{1.034 ± 0.000} & 98.945 ± 0.011         & 1.069 ± 0.001          & 2.160 ± 0.029          & \underline{1.249 ± 0.012}    & \underline{0.320 ± 0.052}    & \underline{2.0 / 10}       & 6.7 / 10               \\
MO-DDOM ($k$=256)                                                   & 1.020 ± 0.001          & \underline{1.476 ± 0.031}    & 1.056 ± 0.001          & \underline{0.552 ± 0.007}    & 1.073 ± 0.004          & 0.808 ± 0.008          & 6.7 / 10            & \underline{3.3 / 10}      \\
\textbf{ManGO} ($k$=256)                           & 1.026 ± 0.002          & 3.373 ± 0.428          & 1.063 ± 0.005          & 0.611 ± 0.026          & 1.248 ± 0.006          & 0.388 ± 0.014          & 4.0 / 10             & 4.7 / 10               \\
\textbf{ $\text{ManGO}^\text{+Self-IS}$} ($k$=256) & \underline{1.028 ± 0.001}    & 2.264 ± 0.237          & \textbf{1.072 ± 0.002} & \textbf{0.498 ± 0.013} & \textbf{1.264 ± 0.002} & \textbf{0.234 ± 0.020} & \textbf{1.3 / 10}   & \textbf{2.0 / 10 }               \\ \bottomrule
\end{tabular}
} 
\label{tab:results_moo_synthetic_avg} 
\end{table*}

  \begin{figure}[t!]
    \centering
    \includegraphics[width= \linewidth]{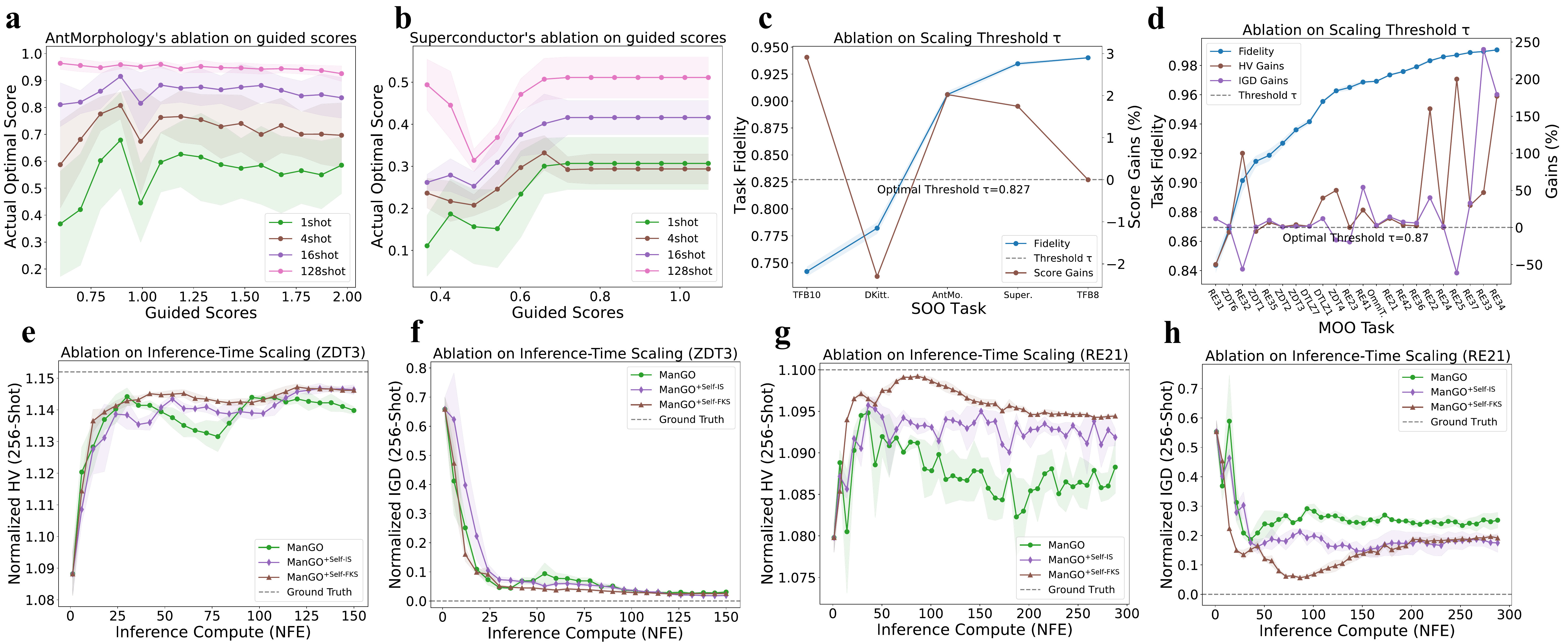}
    \caption{\textbf{Comprehensive ablation studies validating key components of the ManGO framework.}
    (\textbf{a-b}) Flexibility to guidance specification:
    Performance sensitivity to deviations between guided scores and true optima on Ant (a) and Superconductor (b) tasks, demonstrating ManGO's stability under suboptimal guidance conditions.
    (\textbf{c-d}) Fidelity-adaptive scaling: Performance gains relative to baseline fidelity thresholds in offline SOO (c) and MOO (d) tasks, with dashed lines indicating empirically optimal thresholds ($\tau_{\text{opt}} = 0.827$ for SOO; $\tau_{\text{opt}} = 0.87$ for MOO).   
    (\textbf{e-h}) Inference-time scaling efficiency:  HV (e,g) and  IGD (f,h) versus number of function evaluations (NFE) for ZDT3 (e-f) and RE21 (g-h) task  for three approaches, comparing standard denoising, self-IS scaling, and FKS scaling methods.
     Consistent performance improvements are achieved through adaptive noise-space exploration.
}
    \label{fig:ablation_all}
\end{figure}

% \subsection*{Ablation Study}

 \subsection*{Ablation Study on Robustness to Preferred Scores}
 % \subsubsection*{Robustness to Preferred Scores}
Although we use the maximum unobserved score as a guided score (i.e., $\bm{y}_{\rm p}= 1.0$) in Table~\ref{tab:results-designben}, practical scenarios may lack precise knowledge of optimal scores. 
We evaluate ManGO's robustness to suboptimal guided scores by analyzing the best score of generated candidate designs across varying shot numbers.
Figures \ref{fig:ablation_all}.a-b reveal two key insights: 
(i) Increasing the shot number enhances robustness, with stable optimal designs emerging when guided scores exceed $0.7$ (Ant) or $5.5$ (Superconductor) at $128$ shots. 
(ii) Decreasing the shot number to 1, peak performance occurs near but not exactly at  $\bm{y}_{\rm p}$. 
This demonstrates that the generation diversity of diffusion models provides practical robustness for ManGO when $\bm{y}_{\rm p}$ is unknown.

 \subsection*{Ablation Study on Optimal Fidelity Threshold} 
 % \subsubsection*{Optimal Fidelity Threshold for Inference-Time Scaling}   for Inference-Time Scaling
We quantitatively evaluate the design-to-score prediction accuracy through the fidelity metric (Eq.~\ref{eq:fidelity}), which measures the distance between the ground-truth scores and the generated scores of unconditional samples.
Figures~\ref{fig:ablation_all}.c-d show that self-IS-based scaling achieves performance gains on the majority of SOO (optimal fidelity threshold $\tau_{\rm opt}=0.827$) and MOO tasks ($\tau_{\rm opt}=0.87$).
Furthermore, MOO tasks exhibit higher fidelity than SOO tasks due to SOO's higher-dimensional design spaces.
It is more likely to increase performance gains with increasing fidelity values because diffusion models with higher fidelity generate more accurate self-reward signals during inference, enabling more effective noise-space exploration. 

  \subsection*{Ablation Study on Inference-Time Scaling}
% \subsubsection*{Inference-Time Scaling Beyond More Denoising Steps}
We evaluate computation-performance tradeoffs by controlling NFE on ZDT3 in Figures~\ref{fig:ablation_all}.e-f and RE21 in Figures~\ref{fig:ablation_all}.g-h, comparing three approaches: (1) standard guidance with more denoising steps, (2) self-IS-based scaling, and (3) Feynman-Kac-steering (FKS)-based scaling.
For ZDT3, standard guidance achieves competitive performance at $\text{NFE}=31$, demonstrating ManGO's sample efficiency. % i.e., HV >1.14, IGD <0.05,  
However, performance degrades at intermediate NFE before recovering, revealing instability in simple step extension.
In contrast,  scaling-based methods show monotonic improvement with increasing NFE, where FKS scaling temporarily outperforms IS scaling during mid-range NFEs before converging at $\text{NFE}=150$.
The RE21 task exhibits different characteristics: all methods display bell-shaped performance curves, peaking at $\text{NFE}=78$ (FKS), $\text{NFE}=36$ (IS), and $\text{NFE}=36$ (standard).
Scaling-based methods attain higher peak performance over standard guidance while maintaining a sustained advantage after NFE=36.
These results indicate that scaling methods yield superior computational performance compared to simple step extension.
% , namely, better peak performance and more stable overall performance.

\section*{Discussion}

This work introduces ManGO, a framework that fundamentally rethinks offline optimization by learning the underlying design-score manifold through diffusion models. 
Unlike existing approaches that operate in isolated design or score spaces, ManGO's bidirectional modeling unifies forward prediction and backward generation while overcoming OOG challenges. 
The derivative-free guidance mechanism eliminates reliance on error-prone forward models, while the adaptive inference-time scaling dynamically optimizes denoising paths.
Extensive validation across synthetic tasks and real-world applications (robot control, material design, DNA optimization) demonstrates ManGO's consistent superiority among 24 offline-SOO and 10 offline-MOO methods, establishing a new paradigm for data-driven design generation for complex system optimization.

We envision three critical future directions.
First, extending ManGO to high-dimensional and discrete design spaces (e.g., 3D molecular structures) requires developing techniques for learning the latent-based manifold via encoding designs as latents to maintain computational efficiency. 
Recent work on latent diffusion models suggests potential pathways for improvement. 
Second, integrating physics-informed constraints beyond current design-clipping guidance could enhance physical plausibility in domains like metamaterial design, where conservation laws must be preserved. 
Preliminary experiments with physics-informed neural networks show promising results.
Third, developing distributed ManGO variants would enable collaborative optimization across institutions while preserving data privacy, particularly valuable for pharmaceutical development where proprietary molecule datasets exist in isolation.

Several limitations warrant discussion.
 ManGO's current implementation assumes quasi-static system environments, while gradually evolving scenarios would require incremental manifold adaptation mechanisms.
While our adaptive scaling provides partial mitigation, improvement for non-stationary distributions remains an open challenge.
Meanwhile, although ManGO demonstrates strong robustness to preferred scores in identifying optimal designs,  it lacks an iterative refinement mechanism to further improve designs post-generation. 
Recent advances in post-training adaptation of diffusion models, such as controllable fine-tuning and editing, suggest promising pathways to augment ManGO with such capability.

\section*{Methods}

In this section,  we delve into the core components of our approach: (i) training the diffusion model to learn the design-score manifold and (ii) bidirectional guidance generation with the preferred condition.
Finally, we present the training and inference settings of our proposed approach.

\subsection*{Unconditional Training of  Diffusion Model on Score-Augmented Dataset}
To better explore unknown design-score pairs, our method aims to capture the prior probability distribution of the design-score manifold using a diffusion model.
This is a critical step to improve OOG limitation of  backward methods to solve offline optimization problems.
To achieve this, we train an unconditional diffusion model based on a variance-preserving (VP) SDE~\cite{sde_diffusion}. 
The model is trained on a joint design-score dataset $\hat{\mathcal{D}} = \{\hat{\bm{x}_i}: \hat{\bm{x}_i} = (\bm{x}_i, \bm{y}_i) \in \mathbb{R}^{(d+m)}\}_{i=1}^N$, where $\bm{x}_i$ denotes the design and $\bm{y}_i$ represents its corresponding score vector.
Specifically, $\hat{\mathcal{D}}$ is constructed by augmenting the original design dataset $\mathcal{D}_x = \{\bm{x}_i \in \mathbb{R}^{d}\}_{i=1}^N$ with score information. 

Unlike classifier-free diffusion models~\cite{ho2021classifierfree}, which perform conditional training by incorporating score information as a condition with random dropout during training, our method directly learns the joint distribution of designs and scores. 
This eliminates the need to learn the design distribution under varying conditions, instead focusing on capturing the underlying structure of the design-score manifold.
The diffusion process of our manifold-trained model is denoted by the following SDE:
\begin{equation} \label{eq:diffusion_forward}
    d\hat{\bm{x}} = -\frac{1}{2} \beta_t \hat{\bm{x}} \, dt + \sqrt{\beta_t} \, d\bm{w},
\end{equation}
where $\beta_t = \beta_{\min} + (\beta_{\max} - \beta_{\min}) t$ with $t \in [0, 1]$.
The denoising process is represented by the following  reverse-time SDE:
\begin{equation} \label{eq:diffusion_reverse}
    d\hat{\bm{x}} = -\beta_t\left[ \frac{\hat{\bm{x}}}{2} + \nabla_{\hat{\bm{x}}} \log p_t(\hat{\bm{x}}) \right] dt + \sqrt{\beta_t} \, d\tilde{\bm{w}},
\end{equation} 
where $d\tilde{\bm{w}}$ denotes the reverse-time Wiener process. 
Using this process, the pre-trained diffusion model can transform a prior noise point $\hat{\bm{x}}_T \sim p_T(\hat{\bm{x}})$ into a design-score-joint point $\hat{\bm{x}}_0 = (\bm{x}_0, \bm{y}_0)$ on the design-score manifold.

\subsection*{Learning to Reverse Superior Designs via Score-Based Loss Reweighting}
When uniformly sampling data points from the augmented dataset $\hat{\mathcal{D}}$, the diffusion model learns the entire manifold. 
However, this does not fully align with the goal of offline optimization, which aims to find superior or optimal designs.
We introduce a score-based reweighting mechanism into the loss function to steer the model toward regions of the manifold associated with lower scores (indicating better designs).
This mechanism assigns higher weights to samples with lower scores, thereby encouraging the model to prioritize the generation of superior designs and reducing the learning complexity by focusing on the most promising regions of the manifold rather than learning the entire manifold.

Specifically, the weight vector for offline SOO (or MOO) tasks is obtained by applying max-min normalization to the sample scores (or the sample frontier index) based on non-dominated sorting~\cite{deb2002fast} across the entire dataset.
The reweighting mechanism allows the diffusion model to focus on the most promising region of the manifold, which is expressed as:
\begin{equation}\label{eq:weight_vector}
\bm{w}(\hat{\bm{x}} := (\bm{x}, \bm{y})) =
\left\{
\begin{aligned}
  \frac{y_{\max} - y}{y_{\max} - y_{\min}}, &\quad m = 1, \\
  \frac{l_{\text{all}} - l_{\text{NDS}(\bm{y})}}{l_{\text{all}} - 1}, &\quad m > 1,
\end{aligned}
\right.
\end{equation}
where $y_{\min}$ and $y_{\max}$ denote the minimum and maximum scores, respectively, and $l_{\text{all}}$ and $l_{\text{NDS}(\bm{y})}$ represent the total number of frontier layers and the frontier index of $\hat{\bm{x}}$, respectively. 
% Further details about the NDS method are provided in Algorithm~\ref{algo:NDS}. 
Equation (\ref{eq:weight_vector}) indicates that samples with lower scores (or those located on a more advanced frontier) are assigned higher weights in offline SOO (or MOO) tasks.
We note that the normalization-based implementation serves as a foundational step, and future work may explore more efficient reweighting mechanisms. % to further enhance performance

The score function $\nabla_{\hat{\bm{x}}} \log p_t(\hat{\mathbf{x}})$ in Equation (\ref{eq:diffusion_reverse}) is approximated by a time-dependent neural network $\bm{s}_{\bm{\theta}}\left(\hat{\mathbf{x}}_t, t\right)$ with parameters $\bm{\theta}$ based on VP-SDE ~\cite{sde_diffusion}. The network is optimized by minimizing the following loss function:
\begin{equation}\label{eq:diffusion_obj}
\mathcal{L}(\bm{\theta}) = \underset{t}{\mathbb{E}}
\left[
\lambda(t) \underset{\hat{\bm{x}}_0}{\mathbb{E}}
\left[
\bm{w}(\hat{\bm{x}}_0) \underset{\hat{\bm{x}}_t \mid \hat{\bm{x}}_0}{\mathbb{E}}
\left[
\left\|\bm{s}_{\bm{\theta}}\left(\hat{\bm{x}}_t, t \right) - \nabla_{\hat{\bm{x}}} \log p_t\left(\hat{\bm{x}}_t | \hat{\bm{x}}_0\right)\right\|_2^2
\right]
\right]
\right],
\end{equation}
where $\lambda(t)$ is a positive weighting function dependent on $t$, and $ \nabla_{\hat{\bm{x}}} \log p_t\left(\hat{\bm{x}}_t | \hat{\bm{x}}_0\right)$ can be obtained by the diffusion process.

%diffusion sampling
\begin{algorithm}[!t]
   \caption{Self-IS-based ManGO}
   \label{alg:IS-ManGO}
\begin{algorithmic}[1]
    \STATE {\bfseries Input:} Offline dataset $\mathcal{D}$, scaling threshold $\tau$, preferred scores $\bm{y}_{\text{p}}$,
    preferred design constraints $[\bm{x}_{\text{c}_{\text{min}}}, \bm{x}_{\text{c}_{\text{max}}}]$, conditional sample number $K$,  unconditional sample number $M$, duplication size $J$, importance ratio $\alpha_{\rm I}$
    \STATE {\bfseries Output:}  $K$ candidate optimal designs. 
\STATE /* \textit{Phase 1: Training}  */
    \STATE Augment $\mathcal{D}$ and get $\hat{\mathcal{D}} \leftarrow \left\{\hat{\bm{x}}_i: \hat{\bm{x}}_i = (\bm{x}_i, \bm{y}_i)\right\}_{i=1}^N$
    \STATE   Calculate the weight vector $\bm{w}$ with Eq.(\ref{eq:weight_vector})
    \STATE   Training $\bm{s}_{\bm{\theta}}$  with Eq.(\ref{eq:diffusion_obj}) on  $\hat{\mathcal{D}}$
\STATE /*  \textit{Phase 2: Guided generation } */ 
   \STATE Unconditional generation and get  $\mathcal{D}_{\text{uc}}=\{\hat{\bm{x}}_{\text{uc}}^{(i)}\}_{i=1}^M$
    \STATE Compute fidelity $\mathcal{F}(\bm{s}_{\bm{\theta}})$  with Eq.(\ref{eq:fidelity}) on $\mathcal{D}_{\text{uc}}$
     \STATE   Sample random noise $\bm{X}  \leftarrow  \{\hat{\bm{x}}^{(i)}_T  \sim \mathcal{N}(0, \bm{I})\}_{i=1}^K$
    \FOR{$t=T-1$ {\bfseries to} $0$} 
        \IF{$\mathcal{F}(\bm{s}_{\bm{\theta}})>\tau$}
            % \STATE /* Parallel on for $i \in [K]$ samples  */
            \STATE Duplicate $J$ times on $\hat{\bm{x}}^{(i)}_t$ and get  $\bm{X}_i =\{\hat{\bm{x}}^{(i,j)}_t\}_{j=1}^{J}$ for $i \in [K]$ 
            \STATE Guide to reverse $\bm{X}_i \leftarrow \{\hat{\bm{x}}_t^{i,j}\}_{j=1}^{J}$ with Eq.(\ref{eq:guided_generation}) on  $\bm{y}_{\text{p}}$ and $[\bm{x}_{\text{c}_{\text{min}}}, \bm{x}_{\text{c}_{\text{max}}}]$ for $i \in [K]$ 
            \STATE Estimate the denoised state  $\hat{\bm{x}}^{(i,j)}_{0|t}$  with Eq.(\ref{eq:Tweedie_Formula}) for $i, j \in [K], [J]$ 
    \STATE Compute self-supervised reward  $  \mathcal{R}(\hat{\bm{x}}^{(i,j)}_t)  $ with Eq.(\ref{eq:our_reward}) for $i, j \in [K], [J]$ 
            \STATE Compute importance $w^{(i,j)}_t = \exp(\mathcal{R}(\hat{\bm{x}}^{(i,j)}_t)/\alpha_{\rm I})$ for $i, j \in [K], [J]$ 
            \STATE Resample  $j_i^*= \text{Cat} \left(\{w^{(i,j)}_t/\sum_{j=1}^Jw^{(i,j)}_t\}_{j=1}^J \right)$  for $i \in [K]$  
            \STATE  Update $\hat{\bm{x}}^{(i)}_t \leftarrow \hat{\bm{x}}^{(i, j_i^*)}_t $ for $i \in [K]$  and get $\bm{X}  \leftarrow \{\hat{\bm{x}}_t^{(i)}\}_{i=1}^K$
        \ELSE 
            \STATE  Guide to reverse $\bm{X}  \leftarrow \{\hat{\bm{x}}_t^{(i)}\}_{i=1}^K$ with Eq.(\ref{eq:guided_generation}) on  $\bm{y}_{\text{p}}$ and $[\bm{x}_{\text{c}_{\text{min}}}, \bm{x}_{\text{c}_{\text{max}}}]$
    \ENDIF
     \ENDFOR
\end{algorithmic}
\end{algorithm}

\subsection*{Bidirectional Generation on Design-Score Samples via Derivative-Free Guidance}
Diffusion models generate data by iteratively refining random noise into preferred samples through a guided denoising process. 
The pre-trained diffusion model directly generates design-score samples, where each design is paired with a score predicted by the model itself, eliminating the need for an additional surrogate score model. 
This capability allows us to jointly guide the model on both the design and its associated score, enabling the generation of samples that align with our preferences.

% \subsubsection*{Derivative-Free Guidance via Preferred Scores and Preferred Design}
In terms of score-to-design generation, the proposed method enables guided generation of preferred design-score samples through two mechanisms: 
(1) leveraging preferred scores $\bm{y}_{\text{p}}$ to guide the score component of the generated samples, ensuring alignment with desired performance; 
and (2) specifying a design constraint range $[\bm{x}_{\text{c}_{\text{min}}},\bm{x}_{\text{c}_{\text{max}}}]$ to guide the design component, ensuring design feasibility. 
This dual-guidance strategy generates samples that satisfy both design and score constraints without requiring an additional surrogate score model.

Concretely, during the reverse process, the guidance on the score component of the generated sample $\hat{\bm{x}}_t = (\bm{x}_t, \bm{y}_t)$ is formulated as the gradient of the MSE between $\bm{y}_{\text{p}}$ and the current score $\bm{y}_t$, i.e., $\nabla_{\bm{y}_t} \|\bm{y}_{\text{p}} - \bm{y}_t\|_2^2$. 
Similarly, the guidance on the design component is formulated as the gradient of the MSE from the current design $\bm{x}_t$ to the constraint range $[\bm{x}_{\text{c}_{\text{min}}}, \bm{x}_{\text{c}_{\text{max}}}]$, i.e., $\nabla_{\bm{x}_t} \|\bm{x}_t - \text{clip}(\bm{x}_t, \bm{x}_{\text{c}_{\text{min}}}, \bm{x}_{\text{c}_{\text{max}}})\|_2^2$, where $\text{clip}(\bm{x}_t, \bm{x}_{\text{c}_{\text{min}}}, \bm{x}_{\text{c}_{\text{max}}})$ ensures that $\bm{x}_t$ is projected onto the constraint range.

Since both MSE gradients can be computed analytically, our method effectively implements a derivative-free guidance scheme that operates without relying on differentiable models, avoiding the computational overhead of backpropagating (e.g., classifier guidance).
Using the Euler–Maruyama method~\cite{kloeden1992stochastic}, the reverse process can be expressed as:
\begin{equation}\label{eq:guided_generation}
\begin{aligned}
  \hat{\bm{x}}_{t - \Delta t} = \hat{\bm{x}}_t & + \beta_t 
  \left[
     \frac{\hat{\bm{x}}_t}{2}  
     +   \bm{s}_{\bm{\theta}}(\hat{\bm{x}}_t, t)
     + \alpha_x \left(\hat{\bm{x}}_t - \text{clip}(\hat{\bm{x}}_t, \hat{\bm{x}}_{\text{c}_{\text{min}}}, \hat{\bm{x}}_{\text{c}_{\text{max}}})\right)
     + \alpha_y (\hat{\bm{y}}_{\text{p}} - \hat{\bm{y}}_t)
  \right] \Delta t 
  + \sqrt{\beta_t \Delta t} \mathcal{N}(\bm{0}, \bm{I}),
\end{aligned}
\end{equation}
where $\alpha_x$ and $\alpha_y$ control the guidance scale on the design and score components, respectively. Here, $\hat{\bm{x}}_t$, $\hat{\bm{x}}_{\text{c}_{\text{min}}}$, $\hat{\bm{x}}_{\text{c}_{\text{max}}}$, $\hat{\bm{y}}_{\text{p}}$, and $\hat{\bm{y}}_t$ are padded versions of their original vectors with a zero vector $\bm{0}_d$ to match the dimensionality of $\hat{\bm{x}}_t$, like $\hat{\bm{y}}_{\text{p}} = (\bm{0}_d, \bm{y}_{\text{p}})$.

Similarly, the design-to-score prediction  uses the zero-vector-augmented preferred design $\hat{\bm{x}}_{\text{p}}=(\bm{x}_{\rm p}, \bm{0}_{{m}})$ as guidance:
\begin{equation}\label{eq:guided_prediction}
\begin{aligned}
  \hat{\bm{x}}_{t - \Delta t} = \hat{\bm{x}}_t & + \beta_t 
  \left[
     \frac{\hat{\bm{x}}_t}{2}  
     +   \bm{s}_{\bm{\theta}}(\hat{\bm{x}}_t, t)
     + \alpha_x (\hat{\bm{x}}_{\text{p}} - \hat{\bm{x}}_t)
  \right] \Delta t 
  + \sqrt{\beta_t \Delta t} \mathcal{N}(\bm{0}, \bm{I}), 
\end{aligned}
\end{equation}
where the estimated score  of the  preferred design $ {\bm{x}}_{\text{p}}$ is  ${\bm{y}}_0$ from $\hat{\bm{x}}_0=({\bm{x}}_0,{\bm{y}}_0)$.

\subsection*{Achieving Inference-Time Scaling via Self-Supervised Reward}
Diffusion models have been shown to exhibit \textit{inference-time scaling} behavior  \cite{ma2025inferencetimescalingdiffusionmodels,uehara2025rewardguidedcontrolledgenerationinferencetime, singhal2025generalframeworkinferencetimescaling}: their generation quality can be improved by allocating additional computational budgets during inference, even without fine-tuning.
 This stems from a mechanism of noise space exploration, where the pre-trained model optimizes the generation trajectory by adaptively selecting high-reward noise points based on reward feedback.
The reward feedback can be formulated through posterior mean approximation as:
\begin{equation} \label{eq:reward_feedback}
\mathcal{R}(\bm{x}_t) := \mathbb{E}_{\bm{x}_0 \sim p_t^{\text{pre}}(\bm{x}_t)}[r(\bm{x}_0)|\bm{x}_t] = r(\mathbb{E}_{\bm{x}_0 \sim p_t^{\text{pre}}}[\bm{x}_0|\bm{x}_t])  = r(\bm{x}_{0|t}),
\end{equation}
where $\mathcal{R}(\bm{x}_t)$ evaluates the expected reward of samples generated from the noise point $\bm{x}_t$, $p_t^{\text{pre}}$ represents the approximated distribution of the pre-trained model at $t$, $r(\cdot)$ denotes the reward model evaluated at $t=0$, and $\bm{x}_{0|t} = \mathbb{E}_{\bm{x}_0 \sim p_t^{\text{pre}}}[\bm{x}_0|\bm{x}_t]$ denotes that the pre-trained model denoises the noise point from $t$ to 0.
Based on this mechanism, noise points with high rewards are retained for further refinement, while those with low rewards are discarded.

The dual-output capability (design-score generation) of our pre-trained model enables a self-supervised inference-time scaling scheme.
For any intermediate noise point $\hat{\bm{x}}_t$, we estimate the denoised state $\hat{\bm{x}}_{0|t}$ via Tweedie's Formula \cite{ddpm} as:
\begin{equation} \label{eq:Tweedie_Formula}
    \hat{\bm{x}}_{0|t} = (\bm{x}_{0|t}, \bm{y}_{0|t}) = \left( \frac{\bm{x}_{t} + \sqrt{1-\Bar{\alpha}_t}\bm{s}^{(x)}_{\bm{\theta}}(\hat{\bm{x}}_t, t)}{\sqrt{\Bar{\alpha}_t}}, \frac{\bm{y}_{t} + \sqrt{1-\Bar{\alpha}_t}\bm{s}^{(y)}_{\bm{\theta}}(\hat{\bm{x}}_t, t)}{\sqrt{\Bar{\alpha}_t}} \right),
\end{equation}
where $\Bar{\alpha}_t = \exp{[-(\beta_{\max} - \beta_{\min})\frac{t^2}{2} -\beta_{\min}t}]$, and $\bm{s}^{(x)}_{\bm{\theta}}$, $\bm{s}^{(y)}_{\bm{\theta}}$ denote the design and score components of the model output $\bm{s}_{\bm{\theta}}$, respectively.
The estimated score $\bm{y}_{0|t}$ provides self-supervised reward feedback by measuring its deviation from the preferred score:
\begin{equation} \label{eq:our_reward}
    \mathcal{R}(\hat{\bm{x}}_t) =  r(\hat{\bm{x}}_{0|t}) = \| \bm{y}_{\text{p}} - \bm{y}_{0|t} \|_2.
\end{equation}
This eliminates the need for external reward models, enabling the model to autonomously reject low-reward noise points and refine its generation process through self-supervised feedback. 

% \noindent\textbf{Assessing Self-Score Reliability}\
As the effectiveness of inference-time scaling depends on the reward quality \cite{ma2025inferencetimescalingdiffusionmodels}, the effectiveness of our self-supervised reward depends on the fidelity of the pre-trained diffusion model, denoted by $ \mathcal{F}(\bm{s}_{\bm{\theta}})$.
We propose to quantitatively assess $\mathcal{F}(\bm{s}_{\bm{\theta}})$ through the unconditional generation of $\bm{s}_{\bm{\theta}}$. 
Specifically, we first generate a set of unconditional samples $\{{\bm{x}}_{\text{uc}}=({\bm{x}}_{\text{uc}},\bm{y}_{\text{uc}})\}$ by Equation (\ref{eq:guided_generation}) without guidance.
We then filter the samples that are better than training data (i.e.,  ${y}_{\text{uc}}^{(i)} < {y}^{\text{(best)}}_{\text{train}}$ for SOO  and $\bm{y}_{\text{uc}}^{(i)} \prec  \bm{y}^{\text{(best)}}_{\text{train}}$ for MOO) and get the filtered set $\mathcal{D}_{\text{uc}}=\{\hat{\bm{x}}_{\text{uc}}^{(i)}\}_{i=1}^M$.
The fidelity metric $\mathcal{F}(\bm{s}_{\bm{\theta}})$  is subsequently computed as:
\begin{equation} \label{eq:fidelity}
    \mathcal{F}(\bm{s}_{\bm{\theta}})= \exp{(-\frac{1}{M}\sum_{i=1}^M \|\bm{y}_{\text{uc}}^{(i)} - \bm{y}_{\text{train}}^{(j^*)}\|)},
\end{equation}
 where $\mathcal{F}(\bm{s}_{\bm{\theta}}) \in (0,1]$,  $j_i^* = \arg\min_j \|{\bm{x}}_{\text{uc}}^{(i)} - \bm{x}_{\text{train}}^{(j)}\|_2$ denotes the index of the nearest training data to ${\bm{x}}_{\text{uc}}^{(i)}$ in  $\mathcal{D}=\{\hat{\bm{x}}^{(j)}_{\text{train}}=(\bm{x}^{(j)}_{\text{train}},\bm{y}^{(j)}_{\text{train}})\}_{j=1}^{N}$.
A higher value of $\mathcal{F}(\bm{s}_{\bm{\theta}})$ indicates a stronger alignment between the generated and ground-truth scores.
As shown in Figure \ref{fig:ablation_all},  this metric exhibits a strong correlation with inference-time scaling performance based on the self-supervised reward mechanism.
Based on this observation, we implement a conditional logic with $\mathcal{F}(\bm{s}_{\bm{\theta}}) > \tau$ to activate the inference-time scaling scheme during guidance generation.
In this way, we establish a general inference-time scaling framework for offline optimization by substituting external reward models in existing schemes with self-supervised rewards. 

This framework is compatible with various methods, including the IS-based scaling \cite{uehara2025rewardguidedcontrolledgenerationinferencetime} and the FKS-based scaling \cite{singhal2025generalframeworkinferencetimescaling} methods. 
For example, we illustrate the complete process of  ManGO based on  IS with the self-supervised reward in Algorithm \ref{alg:IS-ManGO}.
Self-IS-based ManGO is designed to explore the denoising paths with higher rewards. 
Specifically,  the algorithm first duplicates each noise point $\hat{\bm{x}}_t^{(i)}$ $J$ times.
Each copy then undergoes an independent denoising step as defined in Eq. (\ref{eq:guided_generation}), where the inherent randomness in the reverse process—introduced through the Wiener noise term.
This causes the denoising paths to diverge and results in divergent candidates $\{\hat{\bm{x}}_{t-1}^{(i,j)} \}_{j=1}^J$ at the next timestep.
For each candidate, ManGO estimates its corresponding denoised state to compute self-supervised rewards $\mathcal{R}(\hat{\bm{x}}_{t-1}^{(i,j)})$   Eq.s (\ref{eq:Tweedie_Formula}) and (\ref{eq:our_reward}).
These rewards drive an importance sampling mechanism that prioritizes higher-reward paths, effectively steering the generative process toward more promising regions of the design-score manifold.
This framework leverages the innate stochasticity of diffusion models to explore the noise space dynamically, eliminating the need for explicit external perturbations or reward models. 
Our experiments validate the integration of the IS and FKS approaches with self-supervised rewards.

\subsection*{Training and Inference Settings}

 %~\cite{adam} 
Our model architecture processes three input components: design, scores, and timestep.
 The timestep undergoes standard cosine embedding, while design and score are independently projected via fully connected layers, both to 128-D features. 
These features interact bidirectionally through cross-attention layers, with outputs fused via two multi-layer perceptron (MLP) layers (128 hidden units, Swish activation).
 The fused features are then combined with time embeddings and processed through a three-layer MLP (2048 hidden units, Swish) for reconstruction.
We employ AdamW optimizer with a learning rate (LR) of $5\times 10^{-5}$, a weight decay coefficient of $1 \times 10^{-4}$, and a one-cycle LR scheduler with cosine annealing. 
Training converges in 800 epochs for SOO and 400 epochs for MOO, maintaining original baseline configurations. 
The diffusion process uses  $\beta_{\max}, \beta_{\min} = 1 \times10^{-4}, 5 \times10^{-2}$ for SOO and $ 1 \times10^{-4}, 5 \times10^{-3}$ for MOO.

During inference, we configure $200$ denoising steps for all MOO tasks, Ant and DKittyMorphology of Design-bench tasks,  $5$ for Superconduct tasks, and $250$  for TF-Bind-8 and TF-Bind-10. 
For the guidance scaler,  we set $\alpha_x=1, \alpha_y=1$ for design-constraint trajectory generation in Figure \ref{fig:manifold_soo_moo_row}, and $\alpha_x=0, \alpha_y=1$ for benchmark evaluation.
For the sake of comparison, we disable the fidelity-adaptive activation of the inference-time scaling, denoted as (\ref{eq:fidelity}), in the self-supervised importance sampling-based ManGO.
The scaling methods are activated every five denoising steps, where the IS-based scaling uses beam search with duplication size $J=16$, and the FKS-based scaling uses accumulated maximal rewards.
Complete implementation details are provided in the supplementary material.

\section*{Data Availability}

The datasets used in this paper are publicly available.
The Design-Bench benchmark datasets are available at \url{https://huggingface.co/datasets/beckhamc/design_bench_data}.
The Off-MOO-Bench datasets are available at \url{https://github.com/lamda-bbo/offline-moo}.
The usage of these datasets in this work is permitted under their licenses.
%Source Data are provided with this paper.

\section*{Code Availability}

Codes for this work are available at \url{https://github.com/TailinZhou/ManGO_SOO} for offline SOO tasks and  \url{https://github.com/TailinZhou/ManGO_MOO} for offline MOO tasks.
All experiments and implementation details are thoroughly described in the Experiments section, Methods section, and Supplementary Information.

\section*{Acknowledgements}
The ManGO's main idea and training part were done during Tailin's internship at Huawei Noah's Ark Lab, and the inference part was done at HKUST.
This work was supported in part by the Hong Kong Research Grants Council under the Areas of Excellence scheme grant AoE$/$E-601$/$22-R,  in part by NSFC$/$RGC Collaborative Research Scheme grant CRS$\_$HKUST603$/$22, in part by Guangzhou Municipal Science and Technology Project under Grant 2023A03J0011, Guangdong Provincial Key Laboratory of Integrated Communications, Sensing and Computation for Ubiquitous Internet of Things, and  National Foreign Expert Project, Project Number G2022030026L.
Some icons in Figure \ref{fig:illustration_figure} are from the website \url{https://www.cleanpng.com/}.

\section*{Author Contributions}
T.Z. conceived the idea of this work, implemented the models for experiments, and analyzed the results. 
Z.L.C. and W.L. contributed to the core idea discussion and result interpretation.  
J.Z.  provided critical feedback on the design of inference-time scaling methodology and coordinated the research efforts.
% and J.Z.， Z.T.C., and D.H.K 
All authors participated in manuscript writing and refinement.

\section*{Competing Interests}

The authors declare no competing interests.

\bibliography{sample}

\section*{Supplemental Materials}

 \subsection*{Ablation on duplication size of Self-IS-based ManGO}
\begin{figure}[h]
    \centering
    \includegraphics[width=0.8\linewidth]{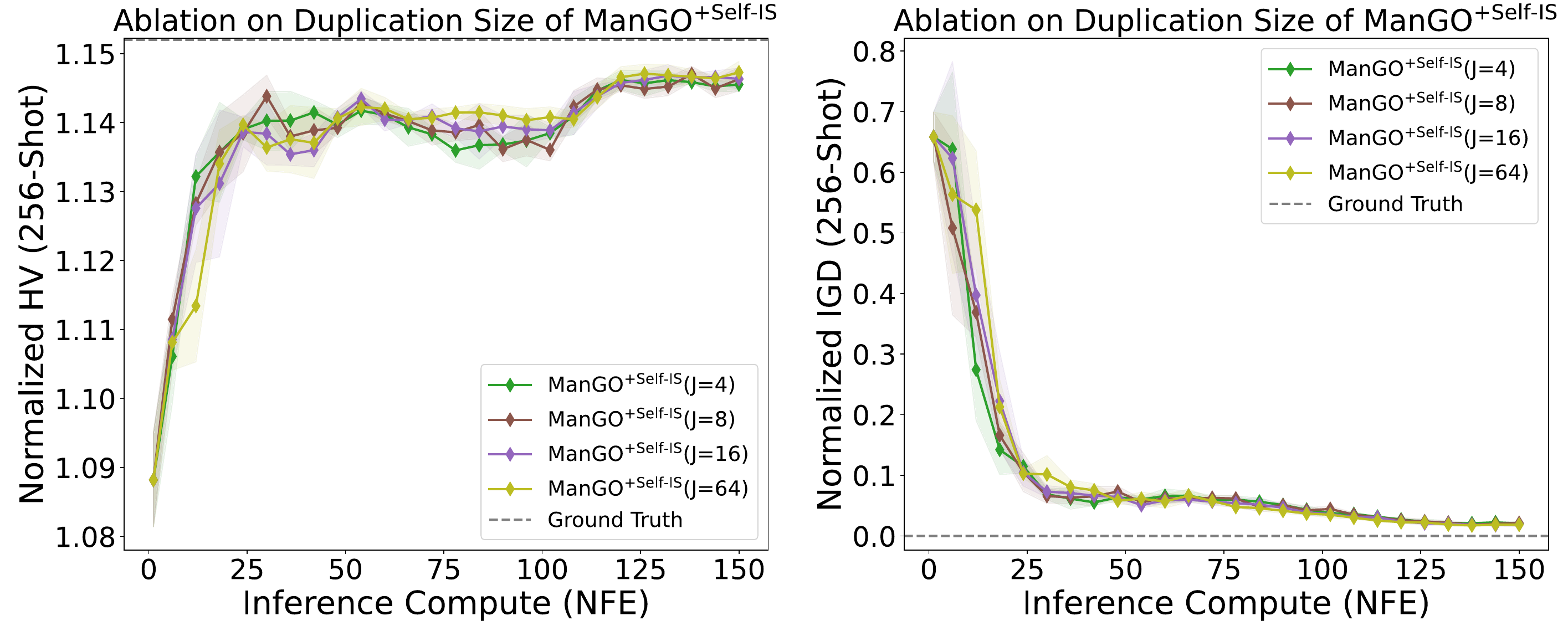}
    \caption{Ablation on duplication size of Self-IS-based ManGO.}
    \label{fig:ablation_scaling_nsmc_J}
\end{figure}

As shown in Figure \ref{fig:ablation_scaling_nsmc_J}, our experiments evaluate duplication size ($J$) settings across $\{4,8,16,64\}$, revealing two key trends:
(i) Performance stability: Larger $J$ values improve performance stability, particularly in the number of function evaluation (NFE) range of 50-100, where HV scores decline.
  Figure \ref{fig:ablation_scaling_nsmc_J} shows that the HV value maintains around 1.14 at $J=64$  and decreases to 1.135 at $J=4$ in this regime, while larger $J$ values introduce better  IGD values.
(ii) Convergence quality: All metrics (HV/IGD) exhibit superior final convergence at NFE=150 with larger $J$.
These results indicate enhanced robustness through more duplicated sampling.
 While $J=64$ delivers optimal results, its 4$\times$ computational overhead versus $J=16$ yields diminishing returns  (only +0.02 HV gain).  Therefore, we select $k=16$ as the default to balance the whole compute costs.

\subsection*{Detailed Experiment Results}
The detailed results are reported in Table \ref{tab:results_moo_synthetic_n_2} for ZDT and OmniTest tasks with $n = 2$ objectives, Table \ref{tab:results_moo_synthetic_n_3} for DTLZ tasks with $n = 3$ objectives, Table \ref{tab:results_moo_re_n_2} for RE tasks with $n = 2$ objectives, Table \ref{tab:results_moo_re_n_3} with $n = 3$ objectives, and Table \ref{tab:results_moo_re_n_4}  with $n = 4$ objectives.
 Note that the results of DTLZ2-DTLZ6 are excluded because of bugs in the original benchmark evaluation
\footnote{\href{https://github.com/lamda-bbo/offline-moo/issues/14}{https://github.com/lamda-bbo/offline-moo/issues/14}}.

\begin{table*}[h]
\centering
\caption{Normalized HV and IGD value of ZDT and OmniTest tasks with $n=2$ objectives in Off-MOO benchmark, where the best and runner-up results on each task are highlighted by \textbf{bold} and \underline{underlined} numbers.}
\resizebox{\linewidth}{!}{
\begin{tabular}{@{}ccccccccccccc@{}}
\toprule
                                                                     & \multicolumn{2}{c}{ZDT1}                                                            & \multicolumn{2}{c}{ZDT2}                                                            & \multicolumn{2}{c}{ZDT3}                                                            & \multicolumn{2}{c}{ZDT4}                                                            & \multicolumn{2}{c}{ZDT6}                                                            & \multicolumn{2}{c}{OmniTest}                                                        \\
                                                                     & HV($\uparrow$)                           & IGD($\downarrow$)                        & HV($\uparrow$)                           & IGD($\downarrow$)                        & HV($\uparrow$)                           & IGD($\downarrow$)                        & HV($\uparrow$)                           & IGD($\downarrow$)                        & HV($\uparrow$)                           & IGD($\downarrow$)                        & HV($\uparrow$)                           & IGD($\downarrow$)                        \\ \midrule
$\mathcal{D}^{\text{(best)}}_{\text{train}}$(Preferred)              & 1.0 (1.164)                              & 1.0 (0.0)                                & 1.0 (1.201)                              & 1.0 (0.0)                                & 1.0 (1.153)                              & 1.0 (0.0)                                & 1.0 (1.023)                              & 1.0 (0.0)                                & 1.0 (1.048)                              & 1.0 (0.0)                                & 1.0 (1.056)                              & 1.0 (0.0)                                \\ \midrule
MM-NSGA2 ($k$=1)                                                     & 0.941 ± 0.0170                           & 1.079 ± 0.0458                           & 0.676 ± 0.0231                           & 1.284 ± 0.0592                           & 0.908 ± 0.0014                           & 0.947 ± 0.0024                           & \textbf{1.031 ± 0.0240}                  & 4.221 ± 0.4111                           & 0.883 ± 0.0009                           & 5.244 ± 0.0733                           & 1.044 ± 0.0102                           & 0.454 ± 0.1290                           \\
MO-DDOM ($k$=1)                                                      & 0.953 ± 0.0073                           & 0.735 ± 0.0026                           & 0.969 ± 0.0030                           & 0.758 ± 0.0146                           & 0.952 ± 0.0041                           & 0.831 ± 0.0179                           & 0.895 ± 0.0195                           & 4.327 ± 0.4032                           & 0.972 ± 0.0127                           & 1.718 ± 0.3118                           & 0.983 ± 0.0020                           & 0.827 ± 0.0458                           \\
\textbf{ManGO} ($k$=1)                              & 1.138 ± 0.0008                           & 0.032 ± 0.0009                           & 1.192 ± 0.0037                           & 0.026 ± 0.0034                           & 1.135 ± 0.0021                           & 0.045 ± 0.0028                           & 0.978 ± 0.0085                           & 3.056 ± 0.0392                           & 1.044 ± 0.0030                           & 0.487 ± 0.2730                           & 1.050 ± 0.0015                           & 0.399 ± 0.0675                           \\
\textbf{$\text{ManGO}^\text{+Self-IS}$} ($k$=1)     & 1.137 ± 0.0021                           & 0.028 ± 0.0014                           & 1.196 ± 0.0036                           & 0.029 ± 0.0027                           & 1.139 ± 0.0006                           & 0.034 ± 0.0042                           & 0.978 ± 0.0164                           & 3.179 ± 0.2338                           & 1.045 ± 0.0022                           & 0.241 ± 0.0833                           & 1.050 ± 0.0014                           & 0.359 ± 0.0169                           \\
\textbf{ $\text{ManGO}^\text{+Self-FKS}$} ($k$=1)   & 1.130 ± 0.0036                           & 0.032 ± 0.0020                           & 1.196 ± 0.0017                           & 0.012 ± 0.0008                           & 1.142 ± 0.0008                           & 0.032 ± 0.0020                           & 0.958 ± 0.0143                           & 3.256 ± 0.2024                           & 1.045 ± 0.0012                           & 0.208 ± 0.0532                           & 1.048 ± 0.0016                           & 0.427 ± 0.0506                           \\ \midrule
MM-MOBO                                                              & 1.037 ± 0.0008                           & 0.649 ± 0.0089                           & 1.064 ± 0.0016                           & 0.676 ± 0.0088                           & 1.033 ± 0.0018                           & 0.677 ± 0.0031                           & 0.959 ± 0.0246                           & 11.192 ± 0.5821                          & 0.724 ± 0.0047                           & 10.420 ± 0.2184                          & \textbf{1.056 ± 0.0000}                  & 0.206 ± 0.0193                           \\
ParetoFlow                                                           & 0.988 ± 0.0060                           & 0.966 ± 0.0552                           & 1.006 ± 0.0028                           & 1.065 ± 0.0549                           & 1.018 ± 0.0082                           & 0.864 ± 0.0457                           & 0.980 ± 0.0164                           & 7.332 ± 0.2448                           & 1.008 ± 0.0061                           & 4.108 ± 1.6264                           & 0.953 ± 0.0573                           & 1.523 ± 0.5666                           \\
MM-NSGA2 ($k$=256)                                                   & \textbf{1.158 ± 0.0003}                  & 0.037 ± 0.0011                           & 1.189 ± 0.0010                           & 0.207 ± 0.0005                           & 1.142 ± 0.0023                           & 0.050 ± 0.0084                           & 0.779 ± 0.0023                           & 15.957 ± 0.0659                          & 1.009 ± 0.0078                           & 1.710 ± 0.1422                           & 1.046 ± 0.0024                           & 1.008 ± 0.0193                           \\
MO-DDOM ($k$=256)                                                    & 0.953 ± 0.0068                           & 0.733 ± 0.0037                           & 0.969 ± 0.0030                           & 0.758 ± 0.0146                           & 0.976 ± 0.0064                           & 0.747 ± 0.0104                           & 0.974 ± 0.0033                           & \textbf{1.622 ± 0.0471}                  & 1.034 ± 0.0121                           & 1.401 ± 0.6451                           & 1.033 ± 0.0009                           & 0.270 ± 0.0096                           \\
\textbf{ManGO} ($k$=256)                            & \underline{1.150 ± 0.0020} & \underline{0.019 ± 0.0014} & 1.199 ± 0.0005                           & 0.006 ± 0.0005                           & 1.139 ± 0.0025                           & 0.035 ± 0.0031                           & 0.998 ± 0.0042                           & \underline{1.953 ± 0.1271} & \textbf{1.047 ± 0.0001}                  & 0.088 ± 0.0187                           & 1.051 ± 0.0020                           & \underline{0.118 ± 0.0145} \\
\textbf{ $\text{ManGO}^\text{+Self-IS}$} ($k$=256)  & 1.142 ± 0.0028                           & \textbf{0.014 ± 0.0006}                  & \textbf{1.201 ± 0.0003}                  & \underline{0.001 ± 0.0003} & \underline{1.144 ± 0.0008} & \textbf{0.021 ± 0.0020}                  & \underline{0.999 ± 0.0044} & 2.115 ± 0.1946                           & 1.044 ± 0.0020                           & \underline{0.072 ± 0.0172} & \underline{1.052 ± 0.0002} & \textbf{0.094 ± 0.0024}                  \\
\textbf{ $\text{ManGO}^\text{+Self-FKS}$} ($k$=256) & 1.133 ± 0.0012                           & 0.021 ± 0.0011                           & \underline{1.200 ± 0.0006} & \textbf{0.000 ± 0.0000}                  & \textbf{1.144 ± 0.0006}                  & \underline{0.025 ± 0.0025} & 0.997 ± 0.0067                           & 2.272 ± 0.0800                           & \underline{1.046 ± 0.0010} & \textbf{0.053 ± 0.0029}                  & 1.051 ± 0.0011                           & 0.201 ± 0.0181                           \\ \bottomrule
\end{tabular}
} 
\label{tab:results_moo_synthetic_n_2}
\end{table*}

\begin{table*}[h]
\centering
\caption{Normalized HV and IGD value of DTLZ tasks with $n=3$ objectives in Off-MOO benchmark, where the best and runner-up results on each task are highlighted by \textbf{bold} and \underline{underlined} numbers. Note that the results of DTLZ2-DTLZ6 are excluded because of bugs in the original benchmark evaluation.}
\resizebox{0.65\linewidth}{!}{
\begin{tabular}{@{}ccccc@{}}
\toprule
                                                                    & \multicolumn{2}{c}{DTLZ1}                         & \multicolumn{2}{c}{DTLZ7}                         \\
                                                                    & HV($\uparrow$)          & IGD($\downarrow$)       & HV($\uparrow$)          & IGD($\downarrow$)       \\ \midrule
$\mathcal{D}^{\text{(best)}}_{\text{train}}$(Preferred)             & 1.0 (1.002)             & 1.0 (0.0)               & 1.0 (1.193)             & 1.0 (0.0)               \\\midrule
MM-NSGA2 ($k$=1)                                                     & 0.974 ± 0.0099          & 0.323 ± 0.0502          & 1.000 ± 0.0142          & 0.948 ± 0.0464          \\
MO-DDOM ($k$=1)                                                      & 0.787 ± 0.0176          & 1.194 ± 0.1069          & 1.077 ± 0.0110          & 0.397 ± 0.0183          \\
\textbf{ManGO} ($k$=1)                              & 0.753 ± 0.0318          & 1.432 ± 0.2049          & 1.189 ± 0.0009          & 0.047 ± 0.0023          \\ 
\textbf{$\text{ManGO}^\text{+Self-IS}$} ($k$=1)     & 0.914 ± 0.0301          & 0.470 ± 0.1672          & \underline{1.192 ± 0.0006}    & 0.030 ± 0.0010          \\
\textbf{ $\text{ManGO}^\text{+Self-FKS}$} ($k$=1)   & 0.931 ± 0.0120          & 0.374 ± 0.0648          & \textbf{1.193 ± 0.0004} & 0.030 ± 0.0010          \\ \midrule
MM-MOBO                                                             & \textbf{1.002 ± 0.0000} & 0.307 ± 0.0182          & 1.147 ± 0.0002          & 0.417 ± 0.0134          \\
ParetoFlow                                                          & 0.999 ± 0.0007          & 0.324 ± 0.1377          & 0.998 ± 0.0166          & 1.020 ± 0.0918          \\
MM-NSGA2 ($k$=256)                                                   & \underline{1.002 ± 0.0001}    & 0.373 ± 0.0308          & 1.170 ± 0.0001          & 1.132 ± 0.0005          \\
MO-DDOM ($k$=256)                                                    & 0.960 ± 0.0117          & \underline{0.220 ± 0.0623}    & 1.149 ± 0.0008          & 0.289 ± 0.0248          \\
\textbf{ManGO} ($k$=256)                            & 0.942 ± 0.0045          & 0.317 ± 0.0255          & 1.189 ± 0.0006          & 0.026 ± 0.0008          \\
\textbf{ $\text{ManGO}^\text{+Self-IS}$} ($k$=256)  & 0.965 ± 0.0062          & 0.236 ± 0.0644          & 1.192 ± 0.0003          & \underline{0.011 ± 0.0010}    \\
\textbf{ $\text{ManGO}^\text{+Self-FKS}$} ($k$=256) & 0.980 ± 0.0031          & \textbf{0.112 ± 0.0166} & 1.192 ± 0.0002          & \textbf{0.006 ± 0.0010} \\ \bottomrule
\end{tabular}
} 
\label{tab:results_moo_synthetic_n_3}
\end{table*}

\begin{table*}[h]
\centering
\caption{Normalized HV and IGD value of RE tasks with $n=2$ objectives in Off-MOO benchmark, where the best and runner-up results on each task are highlighted by \textbf{bold} and \underline{underlined} numbers. All results are normalized by the best HV and IGD in the offline training dataset. The datasets with OOD issues are highlighted in italics.} 
\resizebox{\linewidth}{!}{
\begin{tabular}{@{}ccccccccccc@{}}
\toprule
                                                                     & \multicolumn{2}{c}{\textit{RE21}}                 & \multicolumn{2}{c}{RE22}                          & \multicolumn{2}{c}{RE23}                          & \multicolumn{2}{c}{RE24}                          & \multicolumn{2}{c}{RE25}                          \\
                                                                     & HV($\uparrow$)          & IGD($\downarrow$)       & HV($\uparrow$)          & IGD($\downarrow$)       & HV($\uparrow$)          & IGD($\downarrow$)       & HV($\uparrow$)          & IGD($\downarrow$)       & HV($\uparrow$)          & IGD($\downarrow$)       \\ \midrule
$\mathcal{D}^{\text{(best)}}_{\text{train}}$(Preferred)              & 1.0 (1.100)             & 1.0 (0.0)               & 1.0 (1.006)             & 1.0 (0.0)               & 1.0 (1.019)             & 1.0 (0.0)               & 1.0 (1.053)             & 1.0 (0.0)               & 1.0 (1.010)             & 1.0 (0.0)               \\ \midrule
MM-NSGA2 ($k$=1)                                                     & 0.996 ± 0.0208          & 0.944 ± 0.1898          & \textbf{1.006 ± 0.0000} & 282.500 ± 60.5000       & \underline{1.019 ± 0.0001}    & 0.341 ± 0.1024          & \textbf{1.052 ± 0.0000} & 0.462 ± 0.0010          & 1.005 ± 0.0008          & 0.543 ± 0.0000          \\
MO-DDOM ($k$=1)                                                      & 1.050 ± 0.0012          & 0.636 ± 0.0213          & \textbf{1.006 ± 0.0004} & 7.000 ± 0.5000          & 0.965 ± 0.0507          & 1.434 ± 0.3951          & 1.033 ± 0.0039          & 0.896 ± 0.0201          & 0.997 ± 0.0015          & 1.337 ± 0.1630          \\
\textbf{ManGO} ($k$=1)                              & 1.082 ± 0.0030          & 0.364 ± 0.0139          & 0.986 ± 0.0098          & 31.250 ± 3.2500         & 1.008 ± 0.0003          & 0.961 ± 0.0390          & 1.051 ± 0.0004          & 0.102 ± 0.0157          & 0.993 ± 0.0043          & 1.370 ± 0.3804          \\
\textbf{$\text{ManGO}^\text{+Self-IS}$} ($k$=1)     & 1.089 ± 0.0029          & 0.236 ± 0.0259          & 0.971 ± 0.0100          & 21.000 ± 9.7500         & 1.008 ± 0.0002          & 0.815 ± 0.0439          & 1.051 ± 0.0005          & 0.100 ± 0.0301          & 0.993 ± 0.0060          & 1.435 ± 0.5543          \\
\textbf{ $\text{ManGO}^\text{+Self-FKS}$} ($k$=1)   & 1.089 ± 0.0021          & 0.253 ± 0.0093          & \textbf{1.006 ± 0.0000} & 492.000 ± 0.0000        & 1.009 ± 0.0001          & 0.763 ± 0.0146          & 1.051 ± 0.0003          & 0.114 ± 0.0261          & 1.001 ± 0.0001          & 0.880 ± 0.0163          \\ \midrule
MM-MOBO                                                              & 1.049 ± 0.0113          & 0.481 ± 0.0278          & \textbf{1.006 ± 0.0000} & \underline{5.000 ± 0.2500}    & \underline{1.019 ± 0.0001}    & \textbf{0.146 ± 0.3659} & 1.050 ± 0.0008          & 0.140 ± 0.0204          & \textbf{1.010 ± 0.0000} & \textbf{0.011 ± 0.0000} \\
ParetoFlow                                                           & 1.011 ± 0.0168          & 1.111 ± 0.1056          & \textbf{1.006 ± 0.0000} & 59.250 ± 0.0000         & 1.010 ± 0.0000          & 1.093 ± 0.0000          & 1.047 ± 0.0012          & 0.530 ± 0.1425          & \textbf{1.010 ± 0.0000} & \underline{0.033 ± 0.0109}    \\
MM-NSGA2 ($k$=256)                                                   & 1.087 ± 0.0007          & \textbf{0.074 ± 0.0093} & \textbf{1.006 ± 0.0000} & 492.500 ± 0.0000        & \textbf{1.019 ± 0.0000} & \underline{0.146 ± 0.0439}    & 1.050 ± 0.0002          & 0.482 ± 0.0000          & 1.006 ± 0.0000          & 1.522 ± 0.0000          \\
MO-DDOM ($k$=256)                                                    & 1.049 ± 0.0015          & 0.607 ± 0.0222          & 1.005 ± 0.0000          & \textbf{4.250 ± 0.0000} & 1.010 ± 0.0010          & 1.100 ± 0.0366          & 1.030 ± 0.0038          & 0.912 ± 0.0187          & 1.004 ± 0.0007          & 0.511 ± 0.0761          \\
\textbf{ManGO} ($k$=256)                            & 1.082 ± 0.0026          & 0.276 ± 0.0139          & 0.990 ± 0.0038          & 14.750 ± 2.0000         & 1.009 ± 0.0002          & 0.776 ± 0.0171          & \underline{1.052 ± 0.0001}    & \underline{0.041 ± 0.0057}    & 0.998 ± 0.0015          & 1.022 ± 0.1033          \\
\textbf{ $\text{ManGO}^\text{+Self-IS}$} ($k$=256)  & \textbf{1.092 ± 0.0013} & \underline{0.173 ± 0.0370}    & \textbf{1.006 ± 0.0000} & 9.250 ± 1.000           & 1.008 ± 0.0001          & 0.820 ± 0.0537          & 1.052 ± 0.0003          & \textbf{0.029 ± 0.0030} & 0.997 ± 0.0015          & 1.049 ± 0.0924          \\
\textbf{ $\text{ManGO}^\text{+Self-FKS}$} ($k$=256) & \underline{1.090 ± 0.0003}    & 0.253 ± 0.0037          & \textbf{1.006 ± 0.0000} & 6.500 ± 0.2500          & 1.009 ± 0.0001          & 0.722 ± 0.0171          & \underline{1.052 ± 0.0001}    & 0.102 ± 0.0237          & 1.002 ± 0.0003          & 0.826 ± 0.0054          \\ \bottomrule
\end{tabular}
} 
\label{tab:results_moo_re_n_2} 
\end{table*}

\begin{table*}[h]
\centering
\caption{Normalized HV and IGD value of RE tasks with $n=3$ objectives in Off-MOO benchmark, where the best and runner-up results on each task are highlighted by \textbf{bold} and \underline{underlined} numbers. All results are normalized by the best HV and IGD in the offline training dataset. The datasets with OOD issues are highlighted in italics.} 
\resizebox{\linewidth}{!}{
\begin{tabular}{@{}ccccccccccccccc@{}}
\toprule
                                                                     & \multicolumn{2}{c}{RE31}                          & \multicolumn{2}{c}{RE32}                          & \multicolumn{2}{c}{RE33}                          & \multicolumn{2}{c}{\textit{RE34}}                 & \multicolumn{2}{c}{\textit{RE35}}                 & \multicolumn{2}{c}{\textit{RE36}}                 & \multicolumn{2}{c}{\textit{RE37}}                 \\
                                                                     & HV($\uparrow$)          & IGD($\downarrow$)       & HV($\uparrow$)          & IGD($\downarrow$)       & HV($\uparrow$)          & IGD($\downarrow$)       & HV($\uparrow$)          & IGD($\downarrow$)       & HV($\uparrow$)          & IGD($\downarrow$)       & HV($\uparrow$)          & IGD($\downarrow$)       & HV($\uparrow$)          & IGD($\downarrow$)       \\ \midrule
$\mathcal{D}^{\text{(best)}}_{\text{train}}$(Preferred)              & 1.0 (1.004)             & 1.0 (0.0)               & 1.0 (1.003)             & 1.0 (0.0)               & 1.0 (1.003)             & 1.0 (0.0)               & 1.0 (1.050)             & 1.0 (0.0)               & 1.0 (1.037)             & 1.0 (0.0)               & 1.0 (1.334)             & 1.0 (0.0)               & 1.0 (1.142)             & 1.0 (0.0)               \\ \midrule
MM-NSGA2 ($k$=1)                                                     & 0.994 ± 0.0002          & 0.653 ± 0.1673          & 1.002 ± 0.0006          & 6.046 ± 0.0877          & 0.922 ± 0.0145          & 1.104 ± 0.0026          & 0.695 ± 0.0005          & 7.875 ± 0.1000          & 1.009 ± 0.0057          & 2.162 ± 0.5487          & 1.160 ± 0.0176          & 0.368 ± 0.0418          & 0.761 ± 0.0113          & 2.646 ± 0.1664          \\
MO-DDOM ($k$=1)                                                      & 1.002 ± 0.0010          & \underline{0.523 ± 0.0501}    & \textbf{1.003 ± 0.0002} & \underline{0.654 ± 0.0354}    & 0.985 ± 0.0003          & 1.047 ± 0.0021          & 1.023 ± 0.0011          & 0.502 ± 0.0018          & 1.021 ± 0.0004          & 0.426 ± 0.0138          & 1.213 ± 0.0123          & 0.420 ± 0.0324          & 1.077 ± 0.0002          & 0.479 ± 0.0035          \\
\textbf{ManGO} ($k$=1)                              & 0.996 ± 0.0002          & 0.858 ± 0.0012          & 0.997 ± 0.0023          & 1.703 ± 0.3877          & 0.991 ± 0.0015          & 1.051 ± 0.0034          & 1.005 ± 0.0046          & 0.950 ± 0.0946          & 1.030 ± 0.0010          & 0.242 ± 0.0150          & 1.268 ± 0.0368          & 0.194 ± 0.0506          & 1.069 ± 0.0309          & 0.478 ± 0.0301          \\
\textbf{$\text{ManGO}^\text{+Self-IS}$} ($k$=1)     & 0.998 ± 0.0011          & 0.855 ± 0.0030          & 0.999 ± 0.0001          & 0.977 ± 0.0815          & 0.992 ± 0.0012          & 1.045 ± 0.0006          & 1.026 ± 0.0062          & 0.652 ± 0.0732          & 1.033 ± 0.0005          & 0.137 ± 0.0100          & 1.308 ± 0.0120          & 0.142 ± 0.0106          & 1.108 ± 0.0157          & 0.305 ± 0.0106          \\
\textbf{ $\text{ManGO}^\text{+Self-FKS}$} ($k$=1)   & 0.999 ± 0.0003          & 0.793 ± 0.0983          & 0.999 ± 0.0002          & 1.015 ± 0.0262          & 0.988 ± 0.0035          & 1.036 ± 0.0033          & 1.034 ± 0.0032          & 0.418 ± 0.0304          & \underline{1.033 ± 0.0003}    & 0.135 ± 0.0037          & 1.318 ± 0.0049          & 0.135 ± 0.0060          & \textbf{1.130 ± 0.0007} & 0.202 ± 0.0088          \\ \midrule
MM-MOBO                                                              & 0.984 ± 0.0000          & \textbf{0.341 ± 0.0046} & \textbf{1.003 ± 0.0002} & 2.157 ± 0.5108          & \textbf{1.003 ± 0.0000} & 1.093 ± 0.0008          & 1.008 ± 0.0026          & 1.523 ± 0.1175          & 1.034 ± 0.0001          & 0.955 ± 0.0062          & 1.324 ± 0.0014          & \underline{0.079 ± 0.0097}    & 1.140 ± 0.0000          & 0.238 ± 0.0010          \\
ParetoFlow                                                           & 1.001 ± 0.0028          & 1.023 ± 0.0018          & 1.002 ± 0.0005          & 2.803 ± 0.0015          & 0.992 ± 0.0047          & 1.041 ± 0.0019          & 0.931 ± 0.0000          & 3.816 ± 0.0000          & 0.981 ± 0.0000          & 1.644 ± 0.0000          & 1.186 ± 0.0000          & 1.023 ± 0.0000          & 0.978 ± 0.0223          & 1.290 ± 0.1637          \\
MM-NSGA2 ($k$=256)                                                   & \textbf{1.004 ± 0.0000} & 0.537 ± 0.0110          & \underline{1.002 ± 0.0001}    & 6.508 ± 0.0477          & \underline{0.995 ± 0.0006}    & 1.084 ± 0.0028          & \textbf{1.047 ± 0.0009} & 0.375 ± 0.0768          & 1.012 ± 0.0000          & 6.150 ± 0.0000          & 1.297 ± 0.0003          & 0.076 ± 0.0092          & \textbf{1.126 ± 0.0047} & 0.389 ± 0.0566          \\
MO-DDOM ($k$=256)                                                    & \underline{1.004 ± 0.0002}    & 0.691 ± 0.0116          & \underline{1.002 ± 0.0001}    & \textbf{0.451 ± 0.0123} & 0.990 ± 0.0001          & 1.039 ± 0.0006          & 1.031 ± 0.0006          & 0.507 ± 0.0018          & 1.028 ± 0.0005          & 0.320 ± 0.0050          & 1.252 ± 0.0045          & 0.414 ± 0.0152          & 1.082 ± 0.0008          & 0.444 ± 0.0035          \\
\textbf{ManGO} ($k$=256)                            & 0.998 ± 0.0026          & 0.846 ± 0.0032          & 0.999 ± 0.0003          & 1.008 ± 0.0231          & 0.983 ± 0.0036          & 0.999 ± 0.0429          & 1.013 ± 0.0073          & 0.759 ± 0.0750          & 1.030 ± 0.0008          & 0.195 ± 0.0037          & 1.324 ± 0.0030          & 0.086 ± 0.0161          & 1.091 ± 0.0148          & 0.382 ± 0.0150          \\
\textbf{ $\text{ManGO}^\text{+Self-IS}$} ($k$=256)  & 0.997 ± 0.0002          & 0.828 ± 0.0035          & 1.000 ± 0.0002          & 1.017 ± 0.0185          & 0.991 ± 0.0009          & \textbf{0.987 ± 0.0358} & \underline{1.036 ± 0.0027}    & \textbf{0.327 ± 0.0161} & 1.032 ± 0.0003          & \underline{0.116 ± 0.0025}    & \textbf{1.331 ± 0.0002} & \textbf{0.032 ± 0.0097} & 1.118 ± 0.0078          & \underline{0.179 ± 0.0080}    \\
\textbf{ $\text{ManGO}^\text{+Self-FKS}$} ($k$=256) & 0.997 ± 0.0018          & 0.824 ± 0.0012          & 1.000 ± 0.0002          & 1.015 ± 0.0323          & 0.984 ± 0.0051          & \underline{0.996 ± 0.0435}    & 1.036 ± 0.0037          & \underline{0.350 ± 0.0250}    & \textbf{1.033 ± 0.0001} & \textbf{0.104 ± 0.0050} & \underline{1.325 ± 0.0049}    & 0.106 ± 0.0028          & 1.120 ± 0.0023          & \textbf{0.158 ± 0.0062} \\ \bottomrule
\end{tabular}
} 
\label{tab:results_moo_re_n_3} 
\end{table*}

\begin{table*}[t!]
\centering
\caption{Normalized HV and IGD value of RE tasks with $n=4$ objectives in Off-MOO benchmark, where the best and runner-up results on each task are highlighted by \textbf{bold} and \underline{underlined} numbers. All results are normalized by the best HV and IGD in the offline training dataset. The datasets with OOD issues are highlighted in italics.} 
\resizebox{0.65\linewidth}{!}{
\begin{tabular}{@{}ccccc@{}}
\toprule
                                                                     & \multicolumn{2}{c}{\textit{RE41}}                 & \multicolumn{2}{c}{\textit{RE42}}                 \\
                                                                     & HV($\uparrow$)          & IGD($\downarrow$)       & HV($\uparrow$)          & IGD($\downarrow$)       \\ \midrule
$\mathcal{D}^{\text{(best)}}_{\text{train}}$(Preferred)              & 1.0 (1.116)             & 1.0 (0.0)               & 1.0 (1.505)             & 1.0 (0.0)               \\ \midrule
MM-NSGA2 ($k$=1)                                                     & 0.674 ± 0.0054          & 1.721 ± 0.0734          & 0.885 ± 0.0125          & 1.286 ± 0.0467          \\
MO-DDOM ($k$=1)                                                      & 1.028 ± 0.0005          & 0.870 ± 0.0045          & 1.088 ± 0.0056          & 0.820 ± 0.0106          \\
\textbf{ManGO} ($k$=1)                              & 1.068 ± 0.0061          & 0.568 ± 0.0188          & 1.400 ± 0.0121          & 0.274 ± 0.0173          \\
\textbf{$\text{ManGO}^\text{+Self-IS}$} ($k$=1)     & 1.093 ± 0.0031          & 0.385 ± 0.0195          & 1.386 ± 0.0290          & 0.223 ± 0.0206          \\
\textbf{ $\text{ManGO}^\text{+Self-FKS}$} ($k$=1)   & 1.097 ± 0.0024          & 0.322 ± 0.0136          & 1.409 ± 0.0053          & 0.210 ± 0.0075          \\ \midrule
MM-MOBO                                                              & 1.094 ± 0.0010          & 0.981 ± 0.0305          & 1.151 ± 0.0171          & 0.650 ± 0.0284          \\
ParetoFlow                                                           & 0.894 ± 0.0026          & 1.471 ± 0.0117          & 0.443 ± 0.0000          & 2.302 ± 0.0000          \\
MM-NSGA2 ($k$=256)                                                   & 1.087 ± 0.0041          & 0.429 ± 0.0656          & 1.412 ± 0.0192          & 0.211 ± 0.0379          \\
MO-DDOM ($k$=256)                                                    & 1.038 ± 0.0007          & 0.860 ± 0.0032          & 1.107 ± 0.0072          & 0.757 ± 0.0128          \\
\textbf{ManGO} ($k$=256)                            & 1.081 ± 0.0044          & 0.536 ± 0.0156          & 1.416 ± 0.0073          & 0.239 ± 0.0116          \\
\textbf{ $\text{ManGO}^\text{+Self-IS}$} ($k$=256)  & \underline{1.100 ± 0.0019}    & \underline{0.285 ± 0.0201}    & \textbf{1.429 ± 0.0025} & \textbf{0.184 ± 0.0191} \\
\textbf{ $\text{ManGO}^\text{+Self-FKS}$} ($k$=256) & \textbf{1.103 ± 0.0014} & \textbf{0.267 ± 0.0156} & \underline{1.426 ± 0.0046}    & \underline{0.201 ± 0.0123}    \\ \bottomrule
\end{tabular}
} 
\label{tab:results_moo_re_n_4}\vspace{-1.1em}
\end{table*}

\subsection*{Related Works} \label{sec:related-work}

\subsubsection*{Offline SOO}
The primary methods for offline SOO are the \textit{forward} approach and \textit{backward} approach.
 % \textit{forward} approach} 
The \textit{forward} approach first trains a forward surrogate model $\hat{f}_{ {\theta}}:\mathcal{X} \to \mathbb{R}$, parameterized by $ {\theta}$, to predict the scores of designs, and then 
employs the model as a surrogate evaluator to optimize candidate designs using a search algorithm, e.g., gradient descent.
% Prior works~\cite{coms, nemo,roma, iom, bdi, ict, tri-mentoring, pgs, match-opt, boss} 
Specifically,   the surrogate model is trained by minimizing the mean squared error between the predictions and the true scores as:
\begin{equation*}
    \mathop{\arg\min}_{ {\theta}}\sum\nolimits_{i=1}^N \left( \hat{f}_{ {\theta}} (x_i) - y_i \right)^2/N.
\end{equation*}
With the trained model $\hat{f}_{ {\theta}}$, various ways can obtain the final design, typically gradient descent:
\begin{equation}
    \label{Eq.3}
    x_{t+1}=x_{t} - \eta \left. \nabla_{x} \hat{f}_{ {\theta}}(x) \right|_{x=x_t},~~\text{for }t\in\{0,1,\ldots,T-1\}\end{equation}
where $\eta$ is the search step size, $T$ is the number of steps, and $x_T$ serves as the final design candidate to output.

The \textit{forward} approach has limitations due to its poor performance in out-of-distribution regions. In these areas, the surrogate model $\hat{f}_{ {\theta}}$ may inaccurately overestimate objective scores, misleading the gradient-ascent optimizer into suboptimal regions.
Many recent efforts have been dedicated to addressing this issue, including:

\textit{1) Regularization on the surrogate model: }
Using normalized maximum likelihood, NEMO~\cite{nemo} optimizes the disparity between the surrogate model and the ground-truth function.
RoMA~\cite{roma} improves the smoothness of the model through a combination of pre-training and adaptation, while BOSS~\cite{boss} directly controls the sensitivity of the surrogate model.
GABO~\cite{yao2024generative} introduces adaptive source critic regularization (aSCR) for the surrogate model to constrain the search to reliable areas of the design space. 
In terms of regularization on model predictions, COMs~\cite{coms} penalize identified outliers via a GAN-like procedure~\cite{gan}, whereas  IOM~\cite{iom} maintains representation invariance between the training dataset and design candidate.
Recently, RaM~\cite{tan2024offline} has argued that surrogate models trained with mean squared error do not effectively serve the main goal of offline optimization, which is to select promising designs instead of accurately predicting scores. They propose a ranking-based model that uses learning-to-rank techniques to prioritize designs based on their relative scores.

\textit{2) Employing with ensemble learning: }
An ensemble of surrogate models can lead to significant improvements \cite{design-bench}. In this context, ICT \cite{ict} and Tri-Mentoring \cite{tri-mentoring} each train three symmetric surrogate models and combine them into an ensemble.
ICT employs a semi-supervised learning approach using a pseudo-labeling procedure \cite{ict-semi}, while Tri-Mentoring utilizes a strategy similar to Tri-training \cite{tri-training} but from a pairwise perspective.

\textit{3) Data augmentation: }
Recent works focus on uncovering structural information in datasets for improved learning.
BDI~\cite{bdi} employs both forward and backward mappings to extract knowledge from offline data.
FGM~\cite{fgm} introduces a novel modeling approach that divides the design space into dimension-level cliques for score approximation. 
PGS~\cite{pgs} and Match-OPT~\cite{match-opt} create trajectories from data, with PGS using offline reinforcement learning to predict the gradient ascent optimizer's step size, while Match-OPT ensures model alignment with ground-truth gradients.
DEMO~\cite{demo} modifies designs generated by gradient ascent through a diffusion prior.

 % \textit{backward} approach} 
 The advent of advanced generative models has recently revitalized the \textit{backward} approach across scientific domains.
 The approach first trains a conditioned generative model $p_{ {\theta}}(\mathbf{x}|y)$ and then samples candidate designs from it conditioned on a low score.
For example, MINs~\cite{mins} trains an inverse mapping using a conditioned GAN-like model~\cite{gan, cgan}, while  CbAS~\cite{cbas, auto-cbas} models it as a zero-sum game via a VAE~\cite{vae}.
LEO~\cite{leo} constructs a latent space through an energy-based model that does not require MCMC sampling.
BONET~\cite{bonet} employs trajectories to simulate a black-box optimizer, effectively training an autoregressive model to generate designs using a heuristic. In contrast,  
 GTG~\cite{gtg} focuses on enhancing the quality of trajectories through local search and then directly produces trajectories using a context-conditioned diffusion model.
DDOM~\cite{ddom} directly parameterizes the inverse mapping with a conditional diffusion model~\cite{ddpm} in the design space, while RGD~\cite{anonymous2024robust} further incorporates the explicit proxy guidance into proxy-free diffusion to improve the condition generation.

\subsubsection*{Offline MOO} 
 
The goal of offline MOO is to identify a set of designs that effectively approximate the PF.
While significant progress has been made in MOO research, existing approaches primarily address specific scenarios, such as online MOO and white-box MOO ~\cite{jiang2023multiobjective, park2023botied, gruver2024protein}. 
These methods, however, are not designed to leverage large-scale offline datasets for efficient learning, which constrains their effectiveness in offline scenarios. 
The offline MOO represents a more practical setting, particularly when online evaluations are expensive or potentially hazardous ~\cite{offline-moo}.

\textit{Multi-objective Evolutionary algorithms} are two common methods to tackle offline MOO problems by using a trained surrogate model as the oracle forward function.
Evolutionary algorithms utilize a population-based search strategy that involves iterative processes of parent selection, reproduction, and survivor selection \cite{deb2002fast, zhang2007moea}. 
These algorithms can approximate the Pareto optimal solutions within one execution, with each solution in the population representing a unique trade-off among the objectives~\cite{moea-book}.
  NSGA-II~\cite{nsgaii} is a typical Pareto dominance-based MOEA, using fast non-dominated sorting for selecting solutions. 
NSGA-III~\cite{deb2013evolutionary} is proposed to handle MOO problems with many objectives (having four or more objectives) by using reference points to assist the selection within non-dominated solutions.

\textit{Multi-objective Bayesian optimization} employs a learned surrogate model, e.g., Gaussian process~\cite{gpml} and neural networks, to identify promising candidates through an acquisition function, with each iteration advancing through sampled queries \cite{daulton2023hypervolume, zhang2020random, qing2023pf}.
Existing MOBO methods mainly fall into the following three types.
Hypervolume-based methods mainly adopt three distinct strategies. 
Hypervolume-driven methods incorporate well-known hypervolume metrics within their acquisition functions~\cite{ehvi1,dgemo,qnehvi,hvkg}. 
Scalarization-based approaches transform the MOO problem into one or more SOO tasks through scalarization techniques~\cite{parego,moeadego,mobors,hvrs}.
Information-theoretic-based schemes select points to enhance knowledge of the Pareto front~\cite{pesmo, pfes, jes, pf2es}.
Additionally, various offline SOO training techniques, such as COMs~\cite{trabucco2021conservative}, ROMA~\cite{yu2021roma}, NEMO~\cite{fu2021offline}, ICT~\cite{yuan2023importance}, Tri-Mentoring~\cite{chen2023parallelmentoring}, GradNorm~\cite{gradnorm}, and PcGrad~\cite{yu2020gradient}, can be utilized to improve the effectiveness of training neural networks.

\textit{Generative model-based approach} explores design generation by using multiple desired objective values as conditions in MOO problems with varying settings.
For the molecule-generation setting: structure-property relationships are integrated into a conditional transformer to facilitate a biased generative process by employing a VAE model to recover semantics and property correlations and modeling weights in the latent space \cite{wang2021multi, wang2022multi}.
For white-box settings: Multi-objective guidance is applied within a diffusion framework by \cite{kong2024diffusion}, but equal weights are used for all objectives, failing to capture the full PF. Diversity is introduced through hand-designed diversity penalties instead of uniform weight vectors by \cite{yao2024proud}.
For online settings: Online multi-objective optimization is investigated within a diffusion framework by \cite{gruver2024protein}, with the acquisition being used to guide sample generation. GFlowNet is used as the acquisition function by \cite{zhu2023sampleefficient}, and multiple objectives are integrated into GFlowNet by \cite{jain2023multi}.
For offline settings: An advanced flow-matching model was recently trained on the entire offline dataset by \cite{yuan2024paretoflow}, with multi-objective predictor guidance being used to approximate the PF.

\subsubsection*{Differences between ManGO and existing works} 
While existing approaches in offline optimization rely either on regularized forward surrogate modeling or conditional backward generation, ManGO represents a paradigm shift through its bidirectional manifold learning framework. Our method fundamentally differs from:
(i) forward approaches (e.g., COMs, RaM) that suffer from error accumulation in OOD regions due to gradient misguidance, and
(ii) backward methods (e.g., MINs, DDOM) that are constrained by rigid score-conditioned sampling in design space.
ManGO's capabilities emerge from five key innovations:
\begin{itemize}
    \item \textit{Manifold-aware training}: Holistic unconditional diffusion modeling of joint design-score distributions, capturing intrinsic data geometry.
    \item  \textit{Surrogate-free OOD generation}: A derivative-free guidance mechanism eliminating dependency on auxiliary forward models.
    \item \textit{Dual-space conditioning}: A flexible joint guidance mechanism on both designs and scores, generalizing classifier-free approaches.
    \item \textit{Adaptive inference scaling}: Using self-reward signals to dynamically optimize denoising, surpassing static step methods.
    \item \textit{Unified optimization framework}: A comprehensive framework handling both SOO and MOO through automatic manifold adaptation.
\end{itemize}

\subsection*{Detailed Settings of ManGO}

\subsubsection*{Training Details}
Our model architecture processes three input components: design, scores, and timestep.
 The timestep undergoes standard cosine embedding, while the design and score are independently projected via fully connected layers, both to 128-D features. 
These features interact bidirectionally through cross-attention layers, with outputs fused via two multi-layer perceptron (MLP) layers (128 hidden units, Swish activation).
 The fused features are then combined with time embeddings and processed through a three-layer MLP (2048 hidden units, Swish) for reconstruction.
We employ the AdamW optimizer with a learning rate (LR) of $5\times 10^{-5}$, a weight decay coefficient of $1 \times 10^{-4}$, and a one-cycle LR scheduler with cosine annealing. 
Training converges in 800 epochs for SOO and 400 epochs for MOO, maintaining original baseline configurations. 
The diffusion process uses  $\beta_{\max}, \beta_{\min} = 1 \times10^{-4}, 5 \times10^{-2}$ for SOO and $ 1 \times10^{-4}, 5 \times10^{-3}$ for MOO, respectively.

\subsubsection*{Inference Details}
During inference, we configure $200$ denoising steps for all MOO tasks, Ant and DKittyMorphology of Design-bench tasks,  $5$ for Superconduct tasks, and $250$  for TF-Bind-8 and TF-Bind-10. 
For the guidance scaler,  we set $\alpha_x=1, \alpha_y=1$ for design-constraint trajectory generation in Figure 2, and $\alpha_x=0, \alpha_y=1$ for benchmark evaluation.
For the sake of comparison, we disable the fidelity-aware adaptive activation of the inference-time scaling in the self-supervised importance sampling (self-IS)-based ManGO.
This conservative evaluation demonstrates our method's base performance.
The IS-based noise search is set to be activated using beam search every five denoising steps, and duplication size is $J=16$.
The FKS-based scaling uses accumulated maximal potential rewards and is also set to be activated every five denoising steps.

\subsection*{Detailed Experimental Settings of Design Benchmark}

\subsubsection*{Dataset Preparation}
Following RaM~\cite{tan2024offline}, we create an OOD dataset by extracting high-scoring designs excluded from the training data in Design-Bench~\cite{design-bench}.
In Design-Bench, the training set consists of the bottom-performing  $x\%$ of the full dataset, where $x=40, 50, 60$.
The remaining $(100-x)\% $ of high-scoring data is used to form the unobserved OOD dataset, and all output scores are normalized based on the maximal score of the unobserved OOD dataset.
We randomly sample $10,000$ instances from this subset for all tasks, accessing the complete dataset via the official repository\footnote{\href{https://github.com/brandontrabucco/design-bench}{https://github.com/brandontrabucco/design-bench}}. 

Consistent with prior works~\cite{ddom, bonet, gtg, leo}, we exclude three tasks from Design-Bench~\cite{design-bench} in our evaluation: Hopper~\cite{gym}, ChEMBL~\cite{chembl}, and synthetic NAS on CIFAR10~\cite{cifar10}.
As discussed in earlier studies and this \href{https://github.com/brandontrabucco/design-bench/issues/8}{issue}, Hopper’s implementation in Design-Bench contains a known bug.
For ChEMBL, we omit it because most methods yield nearly identical results, as demonstrated in~\cite{bonet, ddom}, making comparisons uninformative.
NAS is excluded due to the prohibitive computational cost of rigorous multi-seed evaluation, which exceeds our resources.

\subsubsection*{Baseline Setting in Table 1}

We implement the forward surrogate model as a two-hidden-layer MLP with 2048 units per layer using PyTorch~\cite{pytorch}.
We use MSE as loss function and optimize by Adam with learning rate $\eta=0.001$ and learning-rate decay $\gamma=0.98$ to train the forward model with 200 epochs.
For CbAS~\cite{cbas}, MINs~\cite{mins}, and the surrogate-based baselines in Table 1, we adopt the official implementations from the Design-Bench codebase.
Other offline optimization methods are implemented using their publicly released code, retaining their original hyperparameters. However, for DDOM~\cite{ddom} and BONET~\cite{bonet}, we adjust the evaluation budget $k$ from $256$ to $128$ to align with the evaluation protocol of related works.

\subsection*{Detailed Experimental Settings of Off-MOO Benchmark}
\subsubsection*{Dataset Preparation}
We sample $60,000$ instances for all tasks of  Off-MOO Benchmark, accessing the complete dataset via the official repository\footnote{\href{https://github.com/lamda-bbo/offline-moo}{https://github.com/lamda-bbo/offline-moo}}, and follow the benchmark's suggestions to perform data pruning to $30,000$ instances for ZDT3, ZDT4, ZDT6, and OmniTest.
 Besides, we perform min-max normalization within each objective of each problem because different objectives can have different scales, which may result in an imbalanced model update.

Various widely used synthetic functions in MOO literature are employed to evaluate the algorithms.
Specifically, the following benchmark problems are used: DTLZ1, DTLZ7~\cite{dtlz}, ZDT1-4, ZDT6~\cite{zdt}, and Omni-test~\cite{omnitest}. 
VLMOP1-3~\cite{vlmop} are excluded due to lack of PF data, impossible to calculate IGD value.
The solution spaces for all synthetic problems are continuous.
The detailed problem information, Pareto front shape, and reference point can be found in Table~\ref{table:synthetic}.

We also conduct experiments on seven real-world multi-objective engineering design problems adopted from RE suite~\cite{RE}. 
These problems serve as practical applications in various fields.
The search spaces for the problems are continuous except for RE23, which has a mixed solution space (2 variables as integers and 2 as continuous values). The detailed problem information and reference points can be found in Table~\ref{table:re}.
RE61 is excluded because it has 6 objective functions and takes too long to calculate the HV value, particularly for MOBO.

\begin{table}[h]
\centering
\caption{Problem information and reference point for synthetic functions. }
\begin{tabular}{llllll}
\toprule
Name     & $D$ & $m$ & Type & Pareto Front Shape & Reference Point \\ \midrule
DTLZ1    & 7   & 3 &Continuous  & Linear       &    (558.21, 552.30, 568.36)             \\
DTLZ7    & 10  & 3   &Continuous & Disconnected &          (1.10, 1.10, 33.43)       \\
ZDT1     & 30  & 2   &Continuous & Convex       &         (1.10, 8.58)        \\
ZDT2     & 30  & 2   &Continuous & Concave      &          (1.10, 9.59)       \\
ZDT3     & 30  & 2   &Continuous & Disconnected &        (1.10, 8.74)         \\
ZDT4     & 10  & 2   &Continuous & Convex       &          (1.10, 300.42)       \\
ZDT6     & 10  & 2   &Continuous & Concave      &         (1.07, 10.27)        \\
Omnitest & 2   & 2   &Continuous & Convex       &         (2.40, 2.40)        \\
\bottomrule
\end{tabular}\label{table:synthetic}
\end{table}

\begin{table}[h]
\centering
\caption{Problem information and reference point for RE problems.}
\resizebox{0.9\linewidth}{!}{
\begin{tabular}{lllllp{6cm}}
\toprule
Name      & $D$ & $m$ & Type       & Pareto Front Shape &  Reference Point \\ \midrule
RE21 (Four bar truss design)  & 4            & 2          & Continuous &     Convex &   (3144.44, 0.05)    \\
RE22 (Reinforced concrete beam design)  & 3            & 2          & Mixed &     Mixed &   (829.08, 2407217.25)    \\
RE23 (Pressure vessel design) & 4            & 2          & Mixed      &      Mixed, Disconnected   &(713710.88, 1288669.78)     \\
RE24 (Hatch cover design) & 2            & 2          & Continuous      &      Convex   &(5997.83,    43.7)     \\
RE25 (Coil compression spring design) & 3            & 2          & Mixed      &      Mixed, Disconnected   &(124.79, 10038735.00)     \\
RE31 (Two bar truss design) & 3            & 3          & Continuous      &     Unknown   &(808.85, 6893375.82, 6793450.00)     \\
RE32 (Welded beam design) & 4            & 3          & Continuous      &     Unknown   &(290.66, 16552.46, 388265024.00)     \\
RE33 (Disc brake design)                   &          4    &     3       &      Continuous      &   Unknown   & (8.01,    8.84, 2343.30)       \\
RE34 (Vehicle crashworthiness design)                    &         5     &       3     &        Continuous    &   Unknown &(1702.52, 11.68, 0.26)          \\
RE35 (Speed reducer design)                    &         7     &       3     &        Mixed    &   Unknown &(7050.79, 1696.7,  397.83)          \\
RE36 (Gear train design)                    &         4     &       3     &        Integer    &   Concave, Disconnected &(10.21, 60.00         , 0.97)          \\
RE37 (Rocket injector design)                    &     4         &     3      &      Continuous      &   Unknown   &(0.99, 0.96, 0.99)        \\
RE41 (Car side impact design)                     &  7            &        4    &      Continuous      &      Unknown &(42.65,  4.43, 13.08, 13.45)       \\
RE42 (Conceptual marine design)                     &  6            &        4    &      Continuous      &      Unknown &(-26.39, 19904.90, 28546.79, 14.98)       \\ 
  \bottomrule
\end{tabular}}\label{table:re}
\end{table}

\subsubsection*{Baseline Setting in Table 2 and Table 3}
For offline MOO, existing approaches remain relatively under-explored compared to offline SOO. 

According to the Off-MOO Benchmark, an ensemble of independent objective predictors serving as the surrogate model for evolutionary optimization achieves superior results compared to end-to-end and multi-head alternatives.
That is,  multiple models maintain $m$ independent surrogate models for an $m$-objective problem.
Each model learns an objective function independently.
We implement the independent models for all objectives as a two-hidden-layer MLP with 2048 units per layer using PyTorch~\cite{pytorch}.
We use MSE as loss function and optimize by Adam with learning rate $\eta=0.001$ and learning-rate decay $\gamma=0.98$. The independent models are trained w.r.t. the offline dataset for 200 epochs with a batch size of 32.

After training the surrogate model, various methods can be used to obtain the final solution set. 
Following the default setting of Off-MOO Benchmark, we implement NSGA-II as the search algorithm.
Additionally,  we adapt canonic MOBO by substituting Gaussian Processes with the MM ensemble and employ an HV-based acquisition function, qNEHVI~\cite{daulton2021parallel}, which outperforms scalarization and information-theoretic alternatives \cite{offline-moo}.

For the generative methods-based method, we implement the publicly released code of ParetoFlow~\cite{yuan2024paretoflow}, retaining their original hyperparameters.
Note that we use the method of clipping candidate points to fix the bug in the original code that would output candidate points that did not meet the design constraints in RE tasks.
We also extend DDOM as MO-DDOM through multi-score conditioning and adding MM-based design evaluation based on its publicly released code and original hyperparameters.

\end{document}